%% file: main.tex
\documentclass[10pt,twocolumn,letterpaper]{article}

\usepackage[pagenumbers]{cvpr} % To force page numbers, e.g. for an arXiv version

\usepackage[accsupp]{axessibility}
\usepackage{graphicx}
\usepackage{amsmath}
\usepackage{amssymb}
\usepackage{booktabs}
\usepackage{times}
\usepackage{epsfig}
\usepackage{multirow}
\usepackage{subcaption}
\usepackage{enumitem}
\usepackage{makecell}

\DeclareMathOperator*{\argmin}{arg\,min}
\def\vs{\emph{vs.\ }}
\def\eg{\emph{e.g.\ }}
\def\ie{\emph{i.e.\ }}

\usepackage[pagebackref,breaklinks,colorlinks]{hyperref}

\usepackage[capitalize]{cleveref}
\crefname{section}{Sec.}{Secs.}
\Crefname{section}{Section}{Sections}
\Crefname{table}{Table}{Tables}
\crefname{table}{Tab.}{Tabs.}

\begin{document}

\title{Revisiting Random Channel Pruning for Neural Network Compression}

\author{Yawei Li$^1$ \quad Kamil Adamczewski$^2$ \quad Wen Li$^3$ \quad Shuhang Gu$^4$ \quad Radu Timofte$^1$ \quad Luc Van Gool$^{1, 5}$ \\
$^1$Computer Vision Lab, ETH Z\"urich \quad $^2$MPI-IS \quad $^3$UESTC \quad $^4$USYD \quad $^5$KU Leuven\\
{\tt\small \{yawei.li, radu.timofte, vangool\}@vision.ee.ethz.ch}\\
}
\maketitle

%%%%%%%%% ABSTRACT
\begin{abstract}
   Channel (or 3D filter) pruning serves as an effective way to accelerate the inference of neural networks. There has been a flurry of algorithms that try to solve this practical problem, each being claimed effective in some ways. Yet, a benchmark to compare those algorithms directly is lacking, mainly due to the complexity of the algorithms and some custom settings such as the particular network configuration or training procedure. A fair benchmark is important for the further development of channel pruning.
   
   Meanwhile, recent investigations reveal that the channel configurations discovered by pruning algorithms are at least as important as the pre-trained weights. This gives channel pruning a new role, namely searching the optimal channel configuration. In this paper, we try to determine the channel configuration of the pruned models by random search. The proposed approach provides a new way to compare different methods, namely how well they behave compared with random pruning. We show that this simple strategy works quite well compared with other channel pruning methods. We also show that under this setting, there are surprisingly no clear winners among different channel importance evaluation methods, which then may tilt the research efforts into advanced channel configuration searching methods. Code will be released at \url{https://github.com/ofsoundof/random_channel_pruning}.
\end{abstract}

%%%%%%%%% BODY TEXT
%%%%%%%%% BODY TEXT
\section{Introduction}
\label{sec:introduction}
% ------------------------------------------------------------------------------

Since the advent of deep learning based computer vision solutions, network compression has been at the core of reducing the computational complexity of neural networks, accelerating their inference, and enabling their deployment on resource constrained devices~\cite{hinton2015distilling,zhang2016accelerating,zhu2016trained,he2017channel,ye2018rethinking,liu2019rethinking,li2019learning,wang2019haq,li2021towards}. 
Channel pruning (or structured pruning, filter pruning) is one of the approaches that can achieve the acceleration of convolutional neural networks (CNNs)~\cite{li2020group,liu2019metapruning,li2020dhp,ding2019centripetal,he2018amc}. 

The goal of this paper is to conduct an empirical study on channel pruning procedure that is not paid enough attention to, \ie random channel pruning.
By random pruning, we mean that the pruning ratio of each layer is randomly selected and the channels to be pruned within the layer are determined by some criterion. 
Random pruning is frequently referred as a baseline to show the improvements of the state-of-the-art channel pruning methods~\cite{frankle2018lottery,mittal2018recovering,liu2020joint,molchanov2019importance,liu2020discrimination,ding2021resrep,yu2021auto,li2021mixmix,zhang2021exploration}.
Yet, the power of random pruning is not fully released.
By the rigorous study in this paper, we have several striking findings as follows.
% 
% https://tex.stackexchange.com/questions/179452/labels-and-cross-referencing-in-items
\begin{enumerate}[label=F\arabic*]
    \item \label{finding1} When brought to the same setting under random pruning, the recent proposed channel pruning criteria~\cite{he2019filter,molchanov2019importance,luo2020neural,liebenwein2019provable} performs just comparable with the simple L1 and L2 norm based pruning criteria. 
    \item \label{finding2} Compared with channel pruning algorithms that start with a pre-trained model~\cite{he2018soft,lin2019towards,huang2018data,lin2020hrank,liu2020discrimination,he2019filter,zhuang2018discrimination,luo2017thinet,yu2018nisp,he2018amc,denton2014exploiting} (See results in Table~\ref{tbl:benchmark_pruning_method}), random pruning can find a pruned model with comparable or even superior performances. 
    \item \label{finding3} Even compared with advanced pruning methods that optimize the overall network architecture such placement of pooling layers~\cite{liu2019metapruning} and expansion of available network width~\cite{yu2019autoslim}, random pruning still narrows the performance gap (less than 0.5\% on ImageNet classification). 
    \item \label{finding4} Fine-tuning epochs has a strong influence on the performance of the pruned network. High-performing pruned networks usually comes with prolonged fine-tuning epochs. 
\end{enumerate}

Those findings lead to several implications. 
\textit{First of all}, considering~\ref{finding1}, since L1/L2 based channel pruning could perform as well as the other pruning criteria, by the law of Occam's razor, \textbf{most of the cases, the simple L1 and L2 based pruning criteria can just serve the purpose of channel pruning.}
\textit{Secondly}, combining \ref{finding2} and \ref{finding3}, \textbf{random pruning as a neutral baseline, reveals the fundamental development in the field of network pruning.} For algorithms that rely on the predefined network architecture and pre-trained network weight, we haven't gone far since the advent of network pruning. Beyond that, overall network architecture optimization brings additional benefits. The performance difference of most methods fall into a narrow range of 1\%, which is close to the performance of the original network. This on the one hand shows the characteristic of channel pruning, \ie the performance of the channel pruned network is upper bounded by the original network~\footnote{More discussion in the supplementary.}. On the other hand, it shows the difficulty of the problem, \ie every small improvement comes with huge efforts (mostly computation). 
\textit{Thirdly}, considering~\ref{finding4}, \textbf{for a fair comparison and a long-lasting development of the field, fine-tuning epoch should be standardized.} We encourage researchers in this field to explain in detail the training and fine-tuning protocol especially the number of epochs. As such, computational cost could be kept in mind for both researchers and industrial practitioners.
 
The discussion above leads to the unique role that random pruning could play in channel pruning, \ie serving as a baseline to benchmark different channel pruning methods~\cite{blalock2020state}.
On the one hand, random channel pruning could bring different pruning criteria under the same regime.
As such, the different channel importance estimation methods becomes a meta component which is fit to work with the existing methods.
On the other hand, random pruning can become a baseline for other algorithms. Since the performance of channel pruning algorithms can be influenced by a couple of factors especially the fine-tuning procedure, decoupling the influential factors and neutrally showing costs and benefits helps creating clarity. 
Random channel pruning also simplifies the pruning algorithm. Instead of resorting to sophisticated algorithms such as reinforcement learning~\cite{he2018amc}, evolutionary algorithms~\cite{liu2019metapruning}, and proximal gradient descent~\cite{li2020group}, channel pruning can be simplified to randomly sampling a pool of sub-networks and selecting the best from them. 

In this paper, random pruning is studied in two settings.
In the first setting, the task is to prune a pre-trained network.
In the second setting, a pre-trained network is not needed and the pruning algorithm starts with a randomly initialized network. 
The problem is formulated as an architecture search problem. 
To cope with the searching, the network is reparameterized with an architecture similar to that of the original network.
Since the network is trained and pruned from scratch, the second setting is referred to as `pruning from scratch' in this paper. 
In both cases, random pruning aims at searching the optimal numbers of channels for a compact network, by randomly sampling the space of all possible channel configurations. 
Although being extremely easy, random pruning performs surprisingly well compared to the carefully designed pruning algorithms.
The surprising success of random pruning also call for an optimized sampling method that improves the search efficiency.

In short, the contributions of this paper are as follows.
\begin{enumerate}[nosep,label=\arabic*)]
    \item We present random pruning, a simplified channel pruning method as a strong baseline to benchmark other channel pruning methods. The properties of random pruning are analyzed in this paper.
    \item We formalize the basic concepts in channel pruning and try to analyze the reason why random pruning could lead to results comparable to those of carefully designed algorithms.
    \item We benchmark a number of channel pruning methods, incl. criteria for random pruning, to get a feel for the current status of channel pruning.
\end{enumerate}

\section{Related Works}
\label{sec:related_works}
% ------------------------------------------------------------------------------

Channel pruning methods are one of the primary ways to compress neural networks along with the reduction of number of bits in weights via quantization~\cite{gupta2015deep,courbariaux2016binarized} and low rank approximation~\cite{jaderberg2014speeding,zhang2016accelerating,yu2017compressing,li2019learning}.
The purpose of channel pruning methods is to create a thinner architecture while incurring minimal loss in performance relative to that of the original network. 

The early pruning methods concentrate around, so-called, unstructured pruning which removes single parameters from networks~\cite{han2015deep,molchanov2017variational}. These approaches, though interesting theoretically, are more difficult to implement within current hardware and software settings. Therefore, much recent work has focused on structured pruning where network channels can be removed and the models can be practically compressed and accelerated~\cite{anwar2017structured}. 

The pruning methods fit in different paradigms. Most common pruning approaches rely on pruning the parameters based on the magnitude of the weights, such as L1/L2 norm~\cite{ye2018rethinking}, or more recent median pruning~\cite{he2019filter}. When convolved, weights provide direct way to compute and, after pruning, approximate the feature maps~\cite{luo2017thinet}. Assessing output feature maps, which corresponds to the channels can be an alternative to analyze the importance of the parameters in the network~\cite{lin2020hrank,zhuang2018discrimination}.
Another group of pruning methods which have been developed over a few decades utilize the gradient of the loss function with respect to the weights by means of first-order or second-order Taylor series approximations~\cite{lecun1990optimal,hassibi1993second,molchanov2019importance}. In this line of work, the weights of smaller importance have smaller impact on the loss function and therefore can be removed.
Recent approaches are varied and include assessing channel importance by KL-divergence~\cite{luo2020neural}, simulated annealing~\cite{nayman2019xnas}, importance sampling~\cite{baykal2019sipping}, and learning Dirichlet distribution over parameters~\cite{adamczewski2020dirichlet}.

Recently, pruning methods have intertwined with knowledge distillation where two networks, a large and a small one share output information to produce similar results~\cite{hinton2015distilling,tung2019similarity,li2020group,li2021heterogeneity}. Such approach can be also combined with generative adversarial learning for pruning~\cite{lin2019towards}.

Nevertheless, the issue with these methods is that although they provide the importance score for the weights, they neither indicate how many parameters should be pruned nor provide little justification as to choices of the pruned architecture. However, it is widely considered that some of the pruned architectures can be better than others~\cite{frankle2018lottery}. Our work suggests a random architecture search, a simple, unbiased and general approach to compare most of the pruning methods and allows to find a good architecture given a pre-defined model. 

We also noticed other works that try to compare different methods~\cite{liu2019rethinking,huang2021rethinking}. Yet, this paper is fundamentally different from those works in the aim and the enlightenment from the analysis. The aim of \cite{liu2019rethinking} is to identify the value of network pruning as discovering the network architecture whereas our aim is to propose random pruning as a neutral baseline to compare different pruning methods. The study in \cite{huang2021rethinking} ``guides and motivates the researchers to design more reasonable criteria'' while our study finds out that advanced pruning criteria behaves just comparable with the naive L1/L2 norm and calls for an optimized sampling method for efficient search. More discussion is given in the supplementary.

\section{Definition and Preliminaries}
\label{sec:preliminaries}
% ------------------------------------------------------------------------------

\subsection{Basic concepts and formalization}
\label{subsec:concepts}

Before delving into the details of the random pruning procedure in this paper, a couple of concepts are first introduced in this section. 

\medskip
\noindent \textbf{Definition 1} (Random selection in channel pruning).
As far as random pruning in the network is concerned, the randomness could occur in different ways. \textit{I. Fully random.} The channels to be pruned are fully randomly selected without any constraint across layers. This is often used as a weak baseline~\cite{molchanov2019importance,frankle2018lottery,liu2020joint}. \textit{II. Constrained Random.} The pruning ratio of each layer is determined according to some prior knowledge. The pruned channels within a layer are randomly selected. This is studied in~\cite{mittal2018recovering}. \textit{III. Random channel number selection.} The pruning ratio of each layer is randomly sampled and the filters in a layer are pruned according certain criteria. In this paper, the third case of random pruning is studied.

\medskip
\noindent \textbf{Definition 2} (Channel Configuration Space). The channel configuration space $\mathbb{E}$ of a network is defined as the space that contains all of the possible channel number configurations. Let $c_{l_i}$ be the number of channels in a layer $l_i$, then the number of channel configurations within a layer is $2^{c_{l_i}}-1$ (we need at least one channel in a layer) and the space of all the configurations contains $\prod_i^n 2^{c_{l_i}}-1$ samples, where $n$ is the number of layers in an architecture.

Different configurations in the space have varying model complexity (computation, number of parameters, latency) and accuracy. Channel pruning methods aim at finding a target channel configuration that maximizes the accuracy of the network given a fixed model complexity. The configuration space is very different from the parameter space of a network. In the following, two properties that highly influence the channel pruning algorithms are summarized.

\medskip
\noindent \textbf{Property 1:} The channel configuration space is discrete. Conducting differentiable analysis in this space is impossible. This property constitutes a major challenge for channel pruning and architecture search methods. To conduct a search in the space, reinforcement learning, evolutionary algorithm, and also proximal gradient descent have been utilized~\cite{he2018amc,liu2019metapruning,li2020dhp}.

\begin{figure}
    \centering
    \begin{subfigure}[b]{0.25\textwidth}
        \centering
        \includegraphics[width=\textwidth]{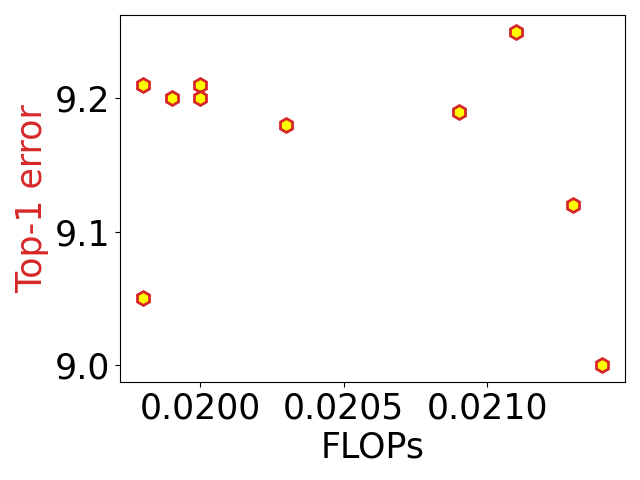}
        % \vspace{-1.7em}
        \caption{Performance of networks in a neighborhood.}
        \label{fig:network_neighborhood}
    \end{subfigure}
    \begin{subfigure}[b]{0.22\textwidth}
        \centering
        \includegraphics[width=\textwidth]{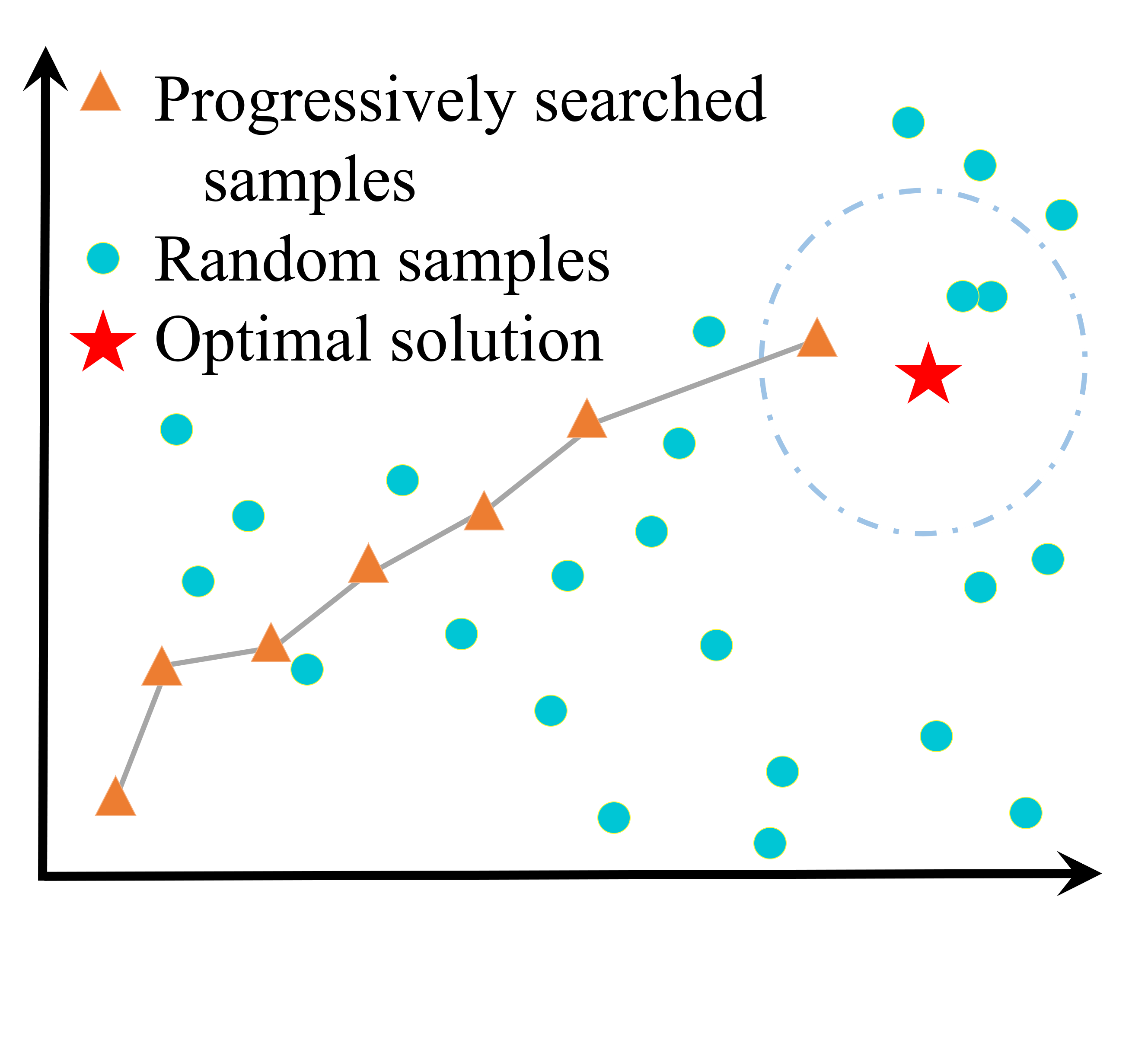}
        % \vspace{-1.7em}
        \caption{Searching of networks in the configuration space.}
        \label{fig:why_random_sample_works}
    \end{subfigure}
    \caption{(a) Slightly modified ResNet20 on CIFAR10 image classification. The accuracy of the networks in the local region of the configuration space does not vary a lot. (b) Random pruning only needs to get a sample in the neighborhood of the optimal solution in the configuration space.}
    \vspace{-8pt}
    \label{fig:teaser}
\end{figure}

\medskip
\noindent \textbf{Property 2:} Slightly changing the channel number of a network does not change the accuracy of the network too much, which means that channel configurations in a local region of the configuration space tend to have similar accuracy. This property is shown in Fig.~\ref{fig:network_neighborhood}, where the accuracy of the network in the top-left region does change a lot.

This property means that the solution to channel pruning problem is not unique. Instead, a group of solutions can exist. This sheds light on the effectiveness of random pruning. Regularization-based methods gradually update from initial networks to the optimal solutions~\cite{ye2018rethinking,li2020group}. By contrast, random pruning only needs to get a sample in the neighborhood of the optimal solution instead of optimal solution itself (See Fig.~\ref{fig:why_random_sample_works}).
As mentioned in the introduction, we study random pruning for channel pruning in two settings. We describe them below.

\medskip
\noindent \textbf{Setting 1: Pruning pre-trained networks}. In this setting, channel pruning methods take a pre-trained network and prune the less important channels according to an importance score.

\medskip
\noindent \textbf{Setting 2: Pruning from scratch}. In this setting, the network is trained from scratch~\cite{yu2018slimmable,yu2019universally,chin2020pareco,berman2020aows,liu2019metapruning,li2020dhp}. During each mini-batch iteration, sub-networks in the allowable channel configuration space in Sec.~\ref{subsec:pruning_space} are trained in parallel such that four sub-networks are sampled and used for parameter update. To cope with the parallel training, a network with architectures similar to the original network is rebuilt according to the description in Sec.~\ref{sec:pruning_scratch}. 
After the training, optimized searching method is used to seek the candidate networks~\cite{yu2019universally,chin2020pareco,berman2020aows}. 
A recent work also incorporates the searching phase into the training phase by penalizing parameters in the rebuilt network, achieving faster convergence~\cite{li2020dhp}.

In the process of pruning the network, the crucial benchmark is the evaluation of the pruned model itself. When pruning and finetuning are done iteratively, 
it is possible to evaluate the performance of the network during pruning. But if the network is severely pruned, the accuracy of the network drops drastically. For example, when directly pruning 30\% of the computation in MobileNetV2, Top-1 error could deteriorate to 90\%.
Directly evaluating the network in this case becomes unreliable. In short, we are faced with the challenge: \textit{how to evaluate the performance of a pruned network in an efficient way?}

For the two pruning settings, there exist different solutions. When pruning a pre-trained network with random pruning, the parameters of the pruned network are updated by minimizing the difference between the feature maps of the pruned network and the original network layer by layer 
Compared with finetuning the network for several epochs, the updating the parameters is more efficient, especially when the number of random samples is large. 
When pruning from scratch, the solution lies in the parallel training procedure of the network. During training, a large number of sub-networks are sampled. The network is trained such that the accuracy of all of the sub-networks tends to decrease. Parallel training arms the network with the capability of interpolating the accuracy of unsampled sub-networks. 
Thus, after training, it is possible to evaluate the performance of the sampled sub-networks reliably.

\subsection{Pruning criteria}
\label{subsec:channel_importance_score}

For channel pruning, it is crucial to evaluate the relative importance of the channels.
There exist several methods that try to measure the channel importance score from different perspectives. The most straightforward method is based on the L1/L2 norms of the filters. 
Consider an individual layer in a network with weight parameters 
$\mathbf{W} = [\mathbf{W}_1, \cdots, \mathbf{W}_n]$, where $\mathbf{W} \in \mathbb{R}^{n \times c \times w \times h}$, $\mathbf{W}_i \in \mathbb{R}^{c \times w \times h}$ denotes the $i$-th output channel of the network (for clarity, we omit the bias). $n$, $c$, $w \times h$ denote the number of output channel, input channel, and kernel size of the layer. Then the L1/L2 norm based importance score is computed as $\mathcal{I}_{norm} = \|\mathbf{W}_i\|_p$, where $p$ could be 1 or 2. 
The filters with smaller norms are likely to be pruned since they generate output feature map with smaller magnitude. 

Yet, some work point out that relying on L1/L2 norms could be problematic since the batch normalization layer could recalibrate the magnitude of the feature map~\cite{ye2018rethinking}. In addition, the "smaller-norm-less-informative" criteria does not respect the distribution of filters in the network~\cite{he2019filter}. Thus, in~\cite{he2019filter} geometric median is proposed to overcome the problem. This criteria discovers the similar filters which could be replaced by the other filters, $\mathcal{I}_{gm} = \sum_j\mathcal{S}(\mathbf{W}_i, \mathbf{W}_j)$, where $\mathcal{S}(\cdot, \cdot)$ denotes the similarity between two filters.

The above criteria are only based on the distribution of the filters in the network, which may not fully respect their influence on the accuracy of the network. Thus, in~\cite{luo2020neural}, Kullback–Leibler divergence is used to measure the importance of a channel by masking out it in the network, $\mathcal{I}_{kl} = \sum_k{P_k \log(\frac{P_k}{Q_k^i})}$, where $P$ is the output probability of the original network and $Q_k^i$ is the probability of the pruned network by masking out the single channel in the network. Channels with smaller KL divergence score have weaker influence on the output probability and can be pruned. However, this method requires to conduct one forward-pass for every channel in the network. This is quite slow compared with other methods. In~\cite{molchanov2019importance}, an accelerated computing method by estimating the prediction error with and without a specific parameter. The estimation is done by taking the first- or second- order Taylor expansion of the prediction error. In short, the importance score of a channel is computed by
$\mathcal{I}_{te} = (\sum_s{g_s w_s})^2$, where $w_s$ denotes a single weight in the channel $\mathbf{W}_i$ and $g_s$ denotes the gradient. Furthermore, in~\cite{baykal2019sipping,liebenwein2019provable}, an empirical sensitivity based on the feature map is proposed. Intuitively, the sensitivity of a feature map reflects the relative impact it has on the pre-activations in the next layer.
In this paper, we try to compare the six metrics under random pruning.

\section{Pruning Pre-trained Networks}
\label{sub:pruning_pretrained}
% ------------------------------------------------------------------------------

\begin{figure}[!t]
    \centering
    \includegraphics[]{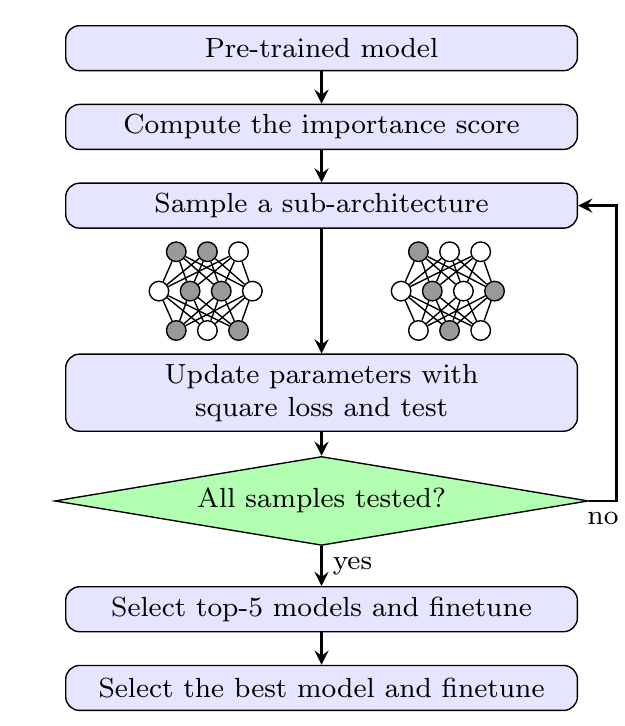}
    \caption{The pipeline of random pruning a pre-trained model.}
    \label{fig:pretrained}
    \vspace{-8pt}
\end{figure}

In this section, the random procedure for pruning a pre-trained model is introduced. The pipeline is shown in Fig.~\ref{fig:pretrained}. The pruning algorithm starts with a pre-trained network. The importance score of individual channels in the pre-trained network is first computed. The importance score is the indicator of which channels should be pruned in the next step. Then we select a number of sub-architectures  and prune the channels with the lowest score. A sub-architecture is formed by sampling pruning ratios for each layer separately, and then pruning the number of channels given by the ratio. A minimum ratio of remaining channels is set. That is, the range for sampling the pruning ratio is $[\eta, 1]$
Next, the parameters of the pruned network are updated by minimizing the squared difference between features maps of the pruned network and the original network, and the accuracy of the pruned network is evaluated on the validation set. The top-5 accurate models are selected and fine-tuned for several epochs to further recover the accuracy of the network. Finally, the model with the best accuracy is selected and fine-tuned for longer epochs. Next, the important steps in the pipeline are explained in detail.

\subsection{Random sampling}
\label{subsec:random_sampling}

The sub-networks are derived by random sampling the pruning ratio for each layer independently.
In total, a population of $N$ sub-networks are sampled. The configurations that meet the target computational complexity are kept. Specifically, let $\mathcal{C}_{prune}$ and $\mathcal{C}_{orig}$ denotes the floating point operations (FLOPs) of the pruned network and the original network, respectively. Then the samples that meet the following criteria are kept, \ie
\begin{equation}
    \left|\frac{\mathcal{C}_{prune}}{\mathcal{C}_{orig}} - \gamma \right| <= \mathcal{T},
    \label{eqn:stop_threshold}
\end{equation}
where $\gamma$ is the overall pruning ratio of the network and $\mathcal{T}$ is the threshold that confines the difference between the actual and target pruning ratio.
During the sampling, the minimum ratio of remaining channels $\eta$ is empirically set around (equal to or slightly smaller than) the overall pruning ratio $\gamma$ based on the following considerations. 
{1)} This setting is simple enough and does not involve  complicated hyper-parameter tuning. {2)} It allows for a reasonably constrained random sampling sub-space for the algorithm to explore.
The setting of $\eta$ prevents the case where a major part of the channels in a layer is pruned. A bottleneck in the network could harm the performance of the pruned network.
The random sampling procedure searches the configuration space. Although it seems to be quite easy, it is shown in the experiments that this procedure is surprisingly competitive.

\subsection{Updating network parameter}

For each sampled sub-architecture, the network is directly pruned according to the per-layer pruning ratio. 
Yet, the accuracy of the network is very likely to drop drastically after pruning, especially when the pruning ratio is high. 
Directly evaluating the pruned network is not reliable. The common practice is to fine-tune the network for a few epochs. But this could be time-consuming considering that a large population of sub-networks are sampled. Instead, we opt for another solution, \ie minimizing the distance between the feature maps of the pruned network and the original network~\cite{he2017channel,luo2017thinet,li2020few}. Let $\mathbf{F}_p \in \mathbb{R}^{n' \times d}$ and $\mathbf{F}_o \in \mathbb{R}^{n \times d}$ denote the feature map of the pruned network and the original network, respectively. Note that the feature maps are reshaped into matrices. Since the network is pruned, its feature map has less channels than the original network, \ie $n' < n$. The parameters in the pruned network is updated by minimizing the following loss function
\begin{equation}
    \mathcal{L} = \argmin_{\mathbf{X}}\|\mathbf{\hat{F}}_o - \mathbf{X}\mathbf{F}_p\|_2^2,
\end{equation}
where $\mathbf{\hat{F}}_o \in \mathbb{R}^{n' \times d}$ is the feature map of the original network with the corresponding channels removed and $\mathbf{X} \in \mathbb{R}^{n' \times n'}$ is the additional parameter that updates the pruned network. The parameter $\mathbf{X}$ can be derived with least square solvers. It can be further merged with the original parameter in the layer of the network. Thus, in fact, no additional parameter or computation is introduced in the pruned network. This parameter updating procedure is done layer-wise.

\section{Pruning From Scratch}
\label{sec:pruning_scratch}
% ------------------------------------------------------------------------------

\begin{figure}[!t]
    \centering
\includegraphics[]{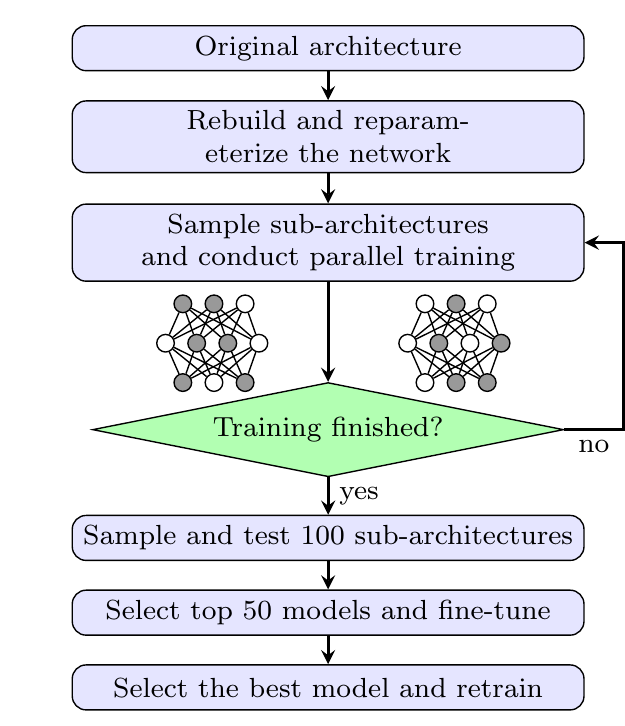}
    \caption{The pipeline of the random pruning from scratch.}
    \label{fig:scratch}
    \vspace{-8pt}
\end{figure}

In this section, the procedure used to prune a network from scratch is described. The pipeline is shown in Fig~\ref{fig:scratch}. In this setting, the pipeline starts with the architecture of the original network. We build a slimmable network according to \cite{yu2018slimmable}. 
The permissible channel configurations are described in Sec.~\ref{subsec:pruning_space}.
Then the network is initialized and trained from scratch. Parallel training is conducted. That is, for each mini-batch iteration during training, four sub-networks are sampled including the complete network and three random samples. Four forward and backward passes are conducted. The gradients during the four backward passes are accumulated and used to update the parameters in the network. The maximum network is always sampled, which guarantees that all of the parameters are updated during one iteration. 
In-place knowledge distillation is used. 

After the training stage, the channel configuration is still searched by random sampling. Thus, a population of $N$ sub-networks satisfying Eqn.~\ref{eqn:stop_threshold} are derived. Owing to the parallel training, the network gains the capability of interpolating the accuracy of unsampled sub-networks. Thus, the sub-networks can be evaluated directly on the validation set and the accuracy is reliable. After that, the top 50 models are further trained for a few epochs.
Finally, the best model among the 50 models is selected and retrained from scratch. In the next subsection, the considerations for rebuilding the network are described.

\subsection{Designing the network pruning space}
\label{subsec:pruning_space}

One problem encountered in this setting is that the total number of sub-networks is quite large. Searching in that large search space is a challenge. Thus, to ease the problem, the pruning space is restricted as follows.
\begin{enumerate}[nosep,label=\arabic*)]
    \item The number of channels is confined to be multiples of 8. Although making the channel number selection discrete, this strategy reduces the possible network samples significantly. For example in the case ResNet-18, the number of possible sub-network configurations is reduced from $2.4 \times 10 ^{24}$ to $2.8 \times 10 ^{14}$. This design is inspired by \textbf{Property 2} of the configuration space. 
    \item The minimum number of channels is reset and rounded to multiples of 8. This again avoids the very narrow bottleneck in the network. For example, when pruning 30\% of the FLOPs of ResNet-18, we empirically require 40\% of the channels must be kept.  
    \item For a fair comparison with pruning pre-trained networks, the maximum network width is not expanded.
\end{enumerate}

\begin{table}[!t]
    \begin{center}
        \scriptsize
        \begin{tabular}{c|c|c|c|c}
            \toprule
            Criterion & \shortstack{Top-1 \\ Error (\%)} & \shortstack{Top-5 \\Error (\%)} & \shortstack{FLOPs [G] / \\ Ratio (\%)} & \shortstack{Params [M] / \\Ratio (\%)} \\ \midrule
            \multicolumn{5}{c}{VGG16, Target FLOPs Ratio 70 \%} \\ \hline
            Baseline	&26.63	&8.5	&15.50	/100.00	&138.4	/100.00	\\
            L1	&27.14	&8.68	&11.11	/71.67	&112.3	/81.14	\\
            L2	&27.01	&8.82	&10.87	/70.10	&128.8	/93.09	\\
            GM	&27.96	&8.77	&11.09	/71.55	&108.4	/78.34	\\
            TE	&27.04	&8.77	&10.65	/68.70	&130.4	/94.27	\\
            ES	&26.76	&8.60	&10.38	/66.98	&130.9	/94.63	\\
            KL	&27.22	&8.88	&10.91	/70.37	&128.9	/93.15	\\ \midrule
            \multicolumn{5}{c}{ResNet18, Target FLOPs Ratio 70 \%} \\ \hline
            Baseline	&30.24	&10.92	&1.82	/100.00	&11.69	/100.00	\\
            L1	&32.08	&12.02	&1.28	/70.53	&8.90	/76.10	\\
            L2	&31.69	&11.79	&1.31	/71.92	&9.58	/81.92	\\
            GM	&31.76	&11.87	&1.30	/71.67	&9.97	/85.27	\\
            TE	&31.76	&11.83	&1.30	/71.38	&9.79	/83.77	\\
            ES	&31.66	&11.86	&1.30	/71.31	&9.41	/80.53	\\
            KL	&31.90	&11.93	&1.31	/72.14	&9.77	/83.54	\\
            Scratch	&31.68	&11.71	&1.28	/70.38	&10.06	/86.05	\\ \midrule
            \multicolumn{5}{c}{ResNet18, Target FLOPs Ratio 50 \%} \\ \hline
            Baseline	&30.24	&10.92	&1.82	/100.00	&11.69	/100.00	\\
            L1	&34.98	&13.81	&0.94	/51.80	&7.49	/64.06	\\
            L2	&35.18	&13.97	&0.93	/51.30	&6.88	/58.88	\\
            GM	&34.50	&13.44	&0.92	/50.57	&8.04	/68.80	\\
            TE	&34.66	&13.81	&0.94	/51.69	&8.02	/68.60	\\
            ES	&35.34	&14.03	&0.94	/51.90	&6.74	/57.67	\\
            KL	&35.10	&13.94	&0.93	/50.88	&7.78	/66.53	\\ \midrule
            \multicolumn{5}{c}{ResNet50, Target FLOPs Ratio 70 \%} \\ \hline
            Baseline	&23.85	&7.13	&4.11	/100.00	&25.56	/100.00	\\
            L1	&24.77	&7.51	&2.81	/68.41	&18.24	/71.35	\\
            L2	&24.33	&7.35	&2.94	/71.48	&20.23	/79.15	\\
            GM	&24.65	&7.40	&2.87	/69.89	&18.60	/72.80	\\
            TE	&24.69	&7.43	&2.89	/70.32	&20.58	/80.53	\\
            ES	&24.66	&7.48	&2.89	/70.26	&18.06	/70.66	\\
            KL	&24.66	&7.49	&2.92	/70.94	&18.60	/72.76	\\ \midrule
            \multicolumn{5}{c}{MobileNetV2, Target FLOPs Ratio 70 \%} \\ \hline
            Baseline	&28.12	&9.71	&0.314	/100.00	&3.50	/100.00	\\
            L1	&32.22	&12.04	&0.224	/71.22	&2.65	/75.74	\\
            L2	&31.84	&11.85	&0.225	/71.63	&2.62	/74.81	\\
            GM	&31.89	&11.88	&0.223	/71.12	&2.69	/76.65	\\
            TE	&32.09	&12.01	&0.223	/70.87	&2.63	/75.16	\\
            ES	&31.93	&11.77	&0.223	/71.03	&2.63	/75.10	\\
            KL	&31.96	&11.93	&0.225	/71.54	&2.64	/75.36	\\
            \bottomrule
        \end{tabular}
    \end{center}
    \vspace{-8pt}
    \caption{Benchmarking channel pruning criteria on ImageNet classification under the scheme of random pruning.}
    \label{tbl:benchmark_imagenet}
    \vspace{-8pt}
\end{table}

%------------------------------------------------------------------------

\section{Experimental Results}
\label{sec:experiments}
% ------------------------------------------------------------------------------

The experimental results are shown in this section. The experiments are conducted on three commonly used networks, including VGG~\cite{simonyan2014very}, ResNet~\cite{he2016deep} and its variants, and MobileNetV2~\cite{sandler2018mobilenetv2}. For ImageNet~\cite{deng2009imagenet} experiments, the pre-trained models provided by PyTorch~\cite{paszke2017automatic} are used as the baseline. 
For CIFAR~\cite{krizhevsky2009learning} experiments, the original network is trained for 300 epochs with the initial learning rate of 0.1 and the batch size of 64. The learning rate decays by 0.1 at the epochs 150 and 225. When pruning the pre-trained models, the pruned architectures with the channels selected by the above methods are tested. The top-5 pruned models are selected and fine-tuned for 5 epochs. 
Eventually to narrow down the search we choose the best model and fine-tune it again to obtain the final pruned model.
For ImageNet and CIFAR, the networks are fine-tuned for 25 and 50 epochs respectively unless otherwise stated. When pruning from scratch, the network is initially trained for 40 epochs. After pruning, the network is reinitialized and retrained for 90 epochs. The population of the sampled sub-network is 100. The threshold $\mathcal{T}$ for random sampling is set to 0.02.

\begin{table}[!t]
    % \small
    % \vspace{-0.1cm}
    \begin{center}
        \scriptsize
        \begin{tabular}{c|c|c|c|c}
            \toprule
            Criterion & \shortstack{Top-1 \\ Error (\%)} & \shortstack{Top-5 \\Error (\%)} & \shortstack{FLOPs [G] / \\ Ratio (\%)} & \shortstack{Params / \\Ratio (\%)} \\ \midrule
            \multicolumn{5}{c}{VGG, CIFAR10} \\ \hline
            Baseline	&5.67	&0.58	&313.80	/100.00	&14.73M	/100.00	\\
            L1	&6.1	&0.69	&160.50	/51.15	&5.05M	/34.32	\\
            L2	&6.06	&0.67	&150.60	/47.99	&6.20M	/42.11	\\
            GM	&5.99	&0.52	&154.60	/49.27	&4.13M	/28.04	\\
            TE	&6.51	&0.61	&157.00	/50.03	&5.84M	/39.63	\\
            ES	&6.21	&0.64	&157.20	/50.10	&7.06M	/47.90	\\
            KL	&6.19	&0.66	&161.50	/51.47	&6.52M	/44.26	\\ \midrule
            \multicolumn{5}{c}{ResNet56, CIFAR10} \\ \hline
            Baseline	&5.58	&0.26	&126.80	/100.00	&855.8k	/100.00	\\
            L1	&6.72	&0.79	&63.60	/50.16	&503.6k	/58.85	\\
            L2	&6.52	&0.76	&64.70	/51.03	&471.4k	/55.08	\\
            GM	&6.39	&0.77	&65.40	/51.58	&504.0k	/58.89	\\
            TE	&6.86	&0.59	&65.70	/51.81	&442.4k	/51.69	\\
            ES	&6.59	&0.67	&65.80	/51.89	&545.6k	/63.75	\\
            KL	&7.12	&0.67	&65.20	/51.42	&443.3k	/51.80	\\ \midrule
            \multicolumn{5}{c}{ResNet20, CIFAR100} \\ \hline
            Baseline	&31.53	&9.87	&41.20	/100.00	&278.3k	/100.00	\\
            L1	&33.41	&10.42	&20.80	/50.49	&176.2k	/63.29	\\
            L2	&33.39	&10.62	&21.00	/50.97	&175.9k	/63.20	\\
            GM	&33.32	&10.35	&20.60	/50.00	&183.8k	/66.03	\\
            TE	&34.24	&10.92	&20.00	/48.54	&168.8k	/60.65	\\
            ES	&33.81	&10.13	&21.00	/50.97	&176.3k	/63.34	\\
            KL	&33.32	&10.62	&21.20	/51.46	&187.5k	/67.35	\\ \bottomrule
        \end{tabular}
    \end{center}
    % \vspace{-8pt}
    \caption{Benchmarking channel pruning criteria on CIFAR10 and CIFAR100 image classification under the scheme of random pruning. More results are given in the supplementary.}
    \label{tbl:benchmark_cifar}
    \vspace{-8pt}
\end{table}

\subsection{Benchmarking channel pruning criteria}
\label{subsec:benchmark_criteria}

It is worth noting that the implemented random pruning method indicates how many channels of each layer should be pruned and \textit{which} channels are pruned is decided by the external criteria.
A range of pruning methods are compared and benchmarked under the scheme of random pruning, including the traditional L1 and L2 norm of the filters (L1, L2), and the recent method based on geometric median (GM)~\cite{he2019filter}, Taylor expansion (TE)~\cite{molchanov2019importance}, KL-divergence importance metric (KL)~\cite{luo2020neural} and empirical sensitivity analysis (ES)~\cite{liebenwein2019provable}. In addition, the method of pruning from scratch based on slimmable networks~\cite{yu2019universally} is also included. 

The benchmark results for ImageNet and CIFAR are shown in Table~\ref{tbl:benchmark_imagenet} and Table~\ref{tbl:benchmark_cifar}, respectively. The FLOP metric is relatively fixed. Since a threshold $\mathcal{T} = 0.02$ is set, the difference between the target overall pruning ratio and the actual overall pruning ratio is within 2\%. During the random sampling, it is difficult to fix both FLOPs and the number of parameters. Thus, the number of parameters of the pruned networks vary. Several conclusions can be drawn by analyzing the results in Table~\ref{tbl:benchmark_imagenet} and Table~\ref{tbl:benchmark_cifar}. \textbf{I.} When comparing different pruning criteria across different networks and datasets under the scheme of random pruning, their performance is close to each other. It is quite surprising that the advanced pruning criteria such as KL and ES do not necessarily outperform the naive ones such as L1 and L2 norm. \textbf{II.} The number of parameters have significant influence on the accuracy of the pruned network. When the computational complexity is about the same, pruned networks with more parameters tend to have lower error rate. \textbf{III.} Considering the above two observations, we conclude that there is no clear winner among the seven compared pruning criteria. \textbf{IV.} Thus, when the pruned networks are fine-tuned for long enough epochs (\eg, more than 25 epochs), the benefits of advanced pruning criteria is substituted by the prolonged training. \textbf{V.} This means that for both pruning a pre-trained model and pruning from scratch, efficient search of the channel configuration space should be at least one of the major research directions.
\begin{table}[!t]
    \begin{center}
        \scriptsize
        \begin{tabular}{c|c|c|c|c|c}
            \toprule
            Methods & Epoch & \shortstack{Top-1 \\ Err. (\%)} & \shortstack{Top-5 \\Err. (\%)} & \shortstack{FLOPs \\ Ratio} & \shortstack{Params\\Ratio} \\ \midrule
            \multicolumn{6}{c}{ResNet50, ImageNet} \\ \hline
            SFP~\cite{he2018soft}	&100	&25.39	&7.94	&58.2	& --	\\
            GAL-0.5~\cite{lin2019towards}	& 30	&28.05	&9.06	&56.97	& 83.14	\\
            SSS~\cite{huang2018data}        &100 &28.18  &9.21   &56.96  &61.15 \\
            HRank~\cite{lin2020hrank} 	& 480	&25.02	&7.67	&56.23	& 63.33	\\
            Random Pruning   & 25   &25.85	&8.01   & 50.72 & 54.99	\\
            Random Pruning   & 75   &25.22	&7.69   & 50.72 & 54.99	\\
            Random Pruning	 & 120	&24.87	&7.48	& 48.99	& 54.12	\\
            AutoPruner~\cite{luo2020autopruner} 	& 32	&25.24	&7.85	&48.79	&--	\\
            Adapt-DCP~\cite{liu2020discrimination}  & 120   &24.85  &7.70   &47.59  &45.01\\
            {FPGM}~\cite{he2019filter}	            & 90	&25.17	&7.68	&46.5	&--	\\
            DCP~\cite{zhuang2018discrimination} 	& 60	&25.05	&7.68	&44.50	&48.44	\\
            ThiNet~\cite{luo2017thinet} 	& 87	&27.97	&9.01	&44.17	& --	\\ \midrule
            {MetaPruning}~\cite{liu2019metapruning}	& 160 & 24.60	& --	&48.78	& --	\\
            {AutoSlim}~\cite{yu2019autoslim}        & 150 & 24.40 & -- & 80.60 & -- \\ \midrule
            
            \multicolumn{6}{c}{MobileNetV2, ImageNet2012} \\ \hline
            MetaPruning~\cite{liu2019metapruning}   & 160 & 28.80 & -- & 72.33 & -- \\
            Random Pruning                          & 120 & 29.10 & -- & 70.87 & -- \\
            AMC~\cite{he2018amc}             & 120 & 29.20 & -- & 70.00 & -- \\
            Adapt-DCP~\cite{liu2020discrimination}  & 310 & 28.55 & -- & 68.92 & -- \\ \midrule
            \multicolumn{6}{c}{ResNet56, CIFAR10} \\ \hline
            GAL-0.5~\cite{lin2019towards}	& 100	&6.62	&--	&63.40	& 88.20	\\
            \cite{li2016pruning} & 40 & 6.94 & -- & 62.40 & 86.30 \\
            NISP~\cite{yu2018nisp} & -- & 6.99 & -- & 56.39 & 57.40 \\
            Random Pruning & 50 & 6.52 & -- & 51.03 & 55.08 \\
            CaP~\cite{minnehan2019cascaded} & -- & 6.78 & -- & 50.20 & -- \\
            ENC~\cite{kim2019efficient} & -- & 7.00 & -- &50.00 & -- \\
            AMC~\cite{he2018amc} & -- & 8.1 & -- &50.00 & -- \\
            Hinge~\cite{li2020group} & 300 & 6.31 & -- & 50.00 &48.73 \\
            KSE~\cite{li2019exploiting} & 200 & 6.77 & --& 48.00 &45.27 \\
            {FPGM}~\cite{he2019filter}	&200	&6.74	&--	&47.4	& --	\\
            {SFP}~\cite{he2018soft}	&300	&6.65	&--	&47.4	& -- \\
            \bottomrule
        \end{tabular}
    \end{center}
    \vspace{-8pt}
    \caption{Benchmarking different channel pruning methods.}
    \label{tbl:benchmark_pruning_method}
    \vspace{-8pt}
\end{table}

\begin{figure*}[!htb]
\begin{subfigure}[t]{0.33\textwidth}
    \centering
    \includegraphics[trim={12 12 11 12},clip,width=0.98\textwidth]{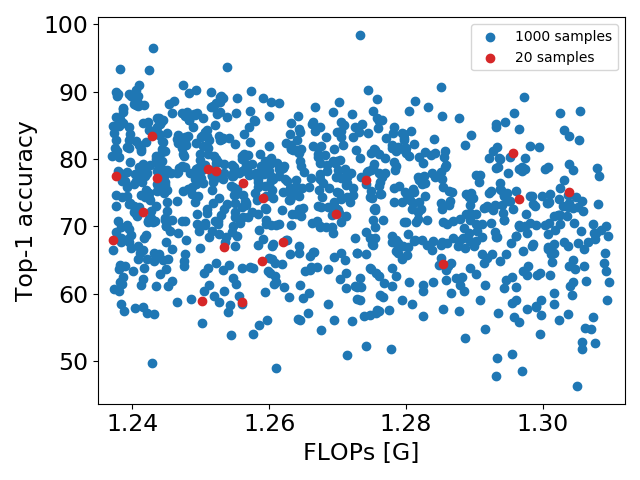}
    \caption{Comparison between 1000 and 20 samples for ResNet18.}
    \label{fig:resnet18_sample_comparison}
\end{subfigure}
\hfill
\begin{subfigure}[t]{0.33\textwidth}
    \centering
    \includegraphics[trim={12 11 12 12},clip,width=0.98\textwidth]{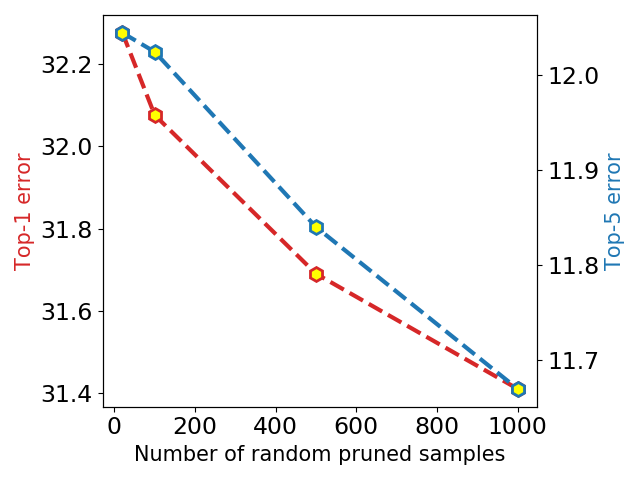}
    \caption{Random samples \vs error rate in ResNet18.}
    \label{fig:resnet18_sample}
\end{subfigure}
\hfill
\begin{subfigure}[t]{0.33\textwidth}
    \centering
    \includegraphics[trim={12 11 12 12},clip,width=0.98\textwidth]{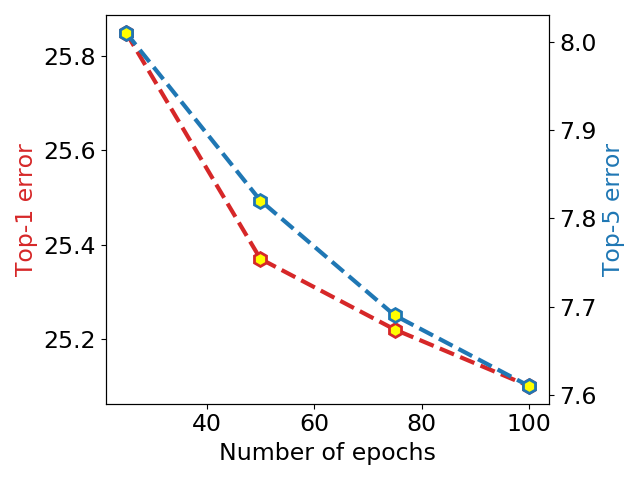}
    \caption{Epochs \vs error rate in ResNet50.}
    \label{fig:resnet50_epoch}
\end{subfigure}
\vspace{-8pt}
\caption{The influence of the random sample size and fine-tuning epochs on the prediction accuracy. Experiments done on ImageNet.}
\label{fig:ablation_study}
\vspace{-10pt}
\end{figure*}

\subsection{Benchmarking channel pruning methods}
\label{subsec:benchmark_method}

To further study the status of channel pruning, we incorporate the results of more methods in Table~\ref{tbl:benchmark_pruning_method}. Two networks are used to benchmark different methods including ResNet50 for ImageNet and ResNet56 for CIFAR10. The results are from the original paper. Note that the number of fine-tuning epochs is also included. This is crucial for comparing different channel pruning methods since the number of training epochs has quite important influence on the accuracy of the final pruned networks. More fine-tuning epochs usually leads to more accurate pruned networks. Ideally, for a fair comparison between different methods, the influence of fine-tuning strategy of the pruning algorithm itself should be decoupled. That is, the number of fine-tuning epochs should be fixed. Yet, this is almost impossible since different methods adopt different training and fine-tuning strategies according to requirements of the methods. In any case, the number of training epochs is still an indicator of the compared algorithms. When the accuracy of two algorithms is close, the one with fewer fine-tuning epochs is obviously better. When the fine-tuning epochs of two algorithms are different, we have a tolerance for the accuracy drop of the one with fewer fine-tuning epochs.

We have a couple conclusions from Table~\ref{tbl:benchmark_pruning_method}. \textbf{I.} On CIFAR10, random pruning performs no worse than any of the compared methods. This shows that for this easy case, random pruning could just serve the purpose. \textbf{II.} On ImageNet, compared with earlier channel pruning methods including SFP~\cite{he2018soft}, GAL~\cite{lin2019towards}, and SSS~\cite{huang2018data}, random pruning outperforms under fewer fine-tuning epochs and severer pruning ratio. \textbf{III.} Random pruning is even comparable with the recent work HRank~\cite{lin2020hrank} considering the longer fine-tuning epochs and larger remained model. \textbf{IV.} Compared with advanced searching methods such as MetaPruning~\cite{liu2019metapruning}, random pruning performs a little bit worse. Yet, we also need to be aware that the slightly changed baseline network for MetaPruning is already in favor of FLOPs reduction. The fine-tuning is also longer. In addition, the potential of random pruning could be fully released as shown in the next subsection. \textbf{V.} Compared with methods that only prune the pre-trained networks, overall architecture optimization such as the placement of pooling layers and expansion of maximum network width could bring additional benefits.

\subsection{Ablation study}

The characteristics of random pruning is ablated.

\noindent \textbf{Influence of the sampling population.} In the former experiments, the number of random samples is fixed to 100. 
In Fig.~\ref{fig:resnet18_sample}, to study the influence of the population size, the number of random samples is increased gradually from 20, 100, 500, to 1000. As expected, the Top-1 and Top-5 error drop steadily from 20 samples to 1000 samples. Meanwhile, the gain of more random sampled does not get saturated. For the studied range, the empirical Top-1 error curve is a monotonically decreasing and convex. This means that the gain of accuracy diminished with increase number of samples. As shown in Fig.~\ref{fig:resnet18_sample_comparison}, when increasing the number of random samples, both better and worse sub-networks could be sampled, which shows the randomness of random pruning. This is acceptable since we are searching for well-performed samples. But from another perspective, this phenomenon also calls for advanced searching methods.

\noindent \textbf{Influence of fine-tuning epochs.} The importance of fine-tuning epochs is already emphasized in Sec.~\ref{subsec:benchmark_method}. Here we quantify the influence of fine-tuning epochs by studying ResNet. The result for ResNet-50 is shown in Fig.~\ref{fig:resnet50_epoch}. The result for ResNet-18 is shown in the supplementary. When the number of fine-tuning epochs is increased from 25 to 100, the Top-1 and Top-5 error of ResNet-50 drops by 0.75\% and 0.4\%, respectively. This shows the significant influence of fine-tuning epochs. Again, when benchmarking, the fine-tuning strategy should be considered.

\noindent \textbf{Analysis of additional computational cost.} Except fine-tuning, other additional computational cost for random pruning includes the evaluation of the pruned models. For pruning pre-trained models, updating the parameters also needs to compute the feature maps, which introduces additional computation. The additional computational cost for evaluation could be reduced by taking out a smaller part (say 5000 for ImageNet) of the validation set for evaluation. This is adopted by some works~\cite{he2018amc}.

\section{Conclusion}
\label{sec:conclusion}
% ------------------------------------------------------------------------------

This work studies the problem of pruning neural network as an unbiased random search for an optimal network architecture. The search can be applied both for learning the architecture from scratch as well as applying it to the pre-trained model with predefined importance score of the channels. As a result, random pruning is a simple, general and explainable baseline which performs well and can be used as a benchmark to more complex pruning methods.

{\footnotesize \noindent \textbf{Acknowledgement:} This work is partially supported by
the ETH Z\"urich Fund (OK), the National Natural Science Foundation of China (Grant No. 62176047), and an Amazon AWS grant.}

%%%%%%%%% REFERENCES
{\small
\bibliographystyle{ieee_fullname}
\bibliography{main}
}

\appendix
\clearpage

\input{supp.tex}

\end{document}

%% file: supp.tex
\maketitle
\begin{center}
\section*{Revisiting Random Channel Pruning for Neural Network Compression: Supplementary Material}
\end{center}

\maketitle

In this supplementary material, we first explain in detail the difference between this work and the previous works. Then we provide a justification of our statement in the main paper ``the performance of the channel pruned network is upper bounded by the original network''. The we show how residual blocks with skip connections  are pruned in Sec.~\ref{sec:pruning_residual}. Finally, more experimental results are given in Sec.~\ref{sec:more_results}.

\section{Difference with Other Works}
In the main paper, we explained the main difference between our work and \cite{liu2019rethinking,huang2021rethinking}. In this supplementary, we provide a detailed comparison between our work and \cite{liu2019rethinking,huang2021rethinking}.

\noindent {\textbf{Difference with~\cite{liu2019rethinking}:}} Our work is different from \cite{liu2019rethinking} in the following aspects.
% \begin{enumerate}[label=\arabic*]
\begin{enumerate}[label=\arabic*)]
    \item \textit{\textbf{Aim.}} The aim of \cite{liu2019rethinking} is to identify the value of network pruning as discovering the network architecture whereas our aim is to propose random pruning as a neutral baseline to compare different pruning methods. 
    
    \item \textit{\textbf{Method.}} How to select the pruning ratio is not thoroughly investigated in \cite{liu2019rethinking} while our work uses random pruning. 
    
    \item \textit{\textbf{Empirical study.}} The empirical study in \cite{liu2019rethinking} is mostly done pairwise by comparing a network resulting from a pruning algorithm and the one trained from scratch. Comparison between different pruning criteria is not done. 
Our work thoroughly compares 6 pruning criteria and 1 architecture search method.
\end{enumerate}

% % 

\noindent{\textbf{Difference with \cite{huang2021rethinking}:}} Our work is different from \cite{huang2021rethinking} in the following aspects.
\begin{enumerate}[label=\arabic*)]
    \item \textit{\textbf{Perspective.}} The analysis in \cite{huang2021rethinking} is conducted on single layers while our work evaluates the overall network performance. 
    \item \textit{\textbf{Conclusion.}} The theoretical and empirical analysis in \cite{huang2021rethinking} mainly support the similarity between norm based pruning criteria. Yet, the empirical study does not support the similarity between importance-based, BN-based, and activation-based pruning criteria. Our study discovers comparable performances between norm-based, importance-based, sensitivity-based, and search based methods. 
    \item \textit{\textbf{Enlightenment.}} The study in \cite{huang2021rethinking} ``guides and motivates the researchers to design more reasonable criteria'' while our study finds out that advanced pruning criteria behaves just comparable with the naive L1/L2 norm ``calls for an optimized sampling method that improves the search efficiency''.
\end{enumerate}

\begin{table}[!t]
    % \small
    % \vspace{-0.1cm}
    \begin{center}
        \scriptsize
        \begin{tabular}{c|c|c|c|c}
            \toprule
            Criterion & \shortstack{Top-1 \\ Error (\%)} & \shortstack{Top-5 \\Error (\%)} & \shortstack{FLOPs [G] / \\ Ratio (\%)} & \shortstack{Params / \\Ratio (\%)} \\ \midrule
            \multicolumn{5}{c}{VGG, CIFAR10} \\ \hline
            Baseline	&5.67	&0.58	&313.80	/100.00	&14.73M	/100.00	\\
            L1	&6.1	&0.69	&160.50	/51.15	&5.05M	/34.32	\\
            L2	&6.06	&0.67	&150.60	/47.99	&6.20M	/42.11	\\
            GM	&5.99	&0.52	&154.60	/49.27	&4.13M	/28.04	\\
            TE	&6.51	&0.61	&157.00	/50.03	&5.84M	/39.63	\\
            ES	&6.21	&0.64	&157.20	/50.10	&7.06M	/47.90	\\
            KL	&6.19	&0.66	&161.50	/51.47	&6.52M	/44.26	\\ \midrule
            \multicolumn{5}{c}{ResNet20, CIFAR10} \\ \hline
            Baseline	&7.48	&0.61	&41.20	/100.00	&272.5k	/100.00	\\
            L1	&9.03	&0.48	&20.90	/50.73	&170.1k	/62.43	\\
            L2	&8.65	&0.55	&20.60	/50.00	&169.9k	/62.37	\\
            GM	&8.69	&0.6	&20.90	/50.73	&188.9k	/69.31	\\
            TE	&8.96	&0.46	&20.90	/50.73	&164.1k	/60.23	\\
            ES	&8.5	&0.63	&25.60	/62.14	&207.8k	/76.24	\\
            KL	&8.77	&0.46	&20.10	/48.79	&165.2k	/60.64	\\\midrule            
            \multicolumn{5}{c}{ResNet56, CIFAR10} \\ \hline
            Baseline	&5.58	&0.26	&126.80	/100.00	&855.8k	/100.00	\\
            L1	&6.72	&0.79	&63.60	/50.16	&503.6k	/58.85	\\
            L2	&6.52	&0.76	&64.70	/51.03	&471.4k	/55.08	\\
            GM	&6.39	&0.77	&65.40	/51.58	&504.0k	/58.89	\\
            TE	&6.86	&0.59	&65.70	/51.81	&442.4k	/51.69	\\
            ES	&6.59	&0.67	&65.80	/51.89	&545.6k	/63.75	\\
            KL	&7.12	&0.67	&65.20	/51.42	&443.3k	/51.80	\\\midrule
            \multicolumn{5}{c}{ResNet20, CIFAR100} \\ \hline
            Baseline	&31.53	&9.87	&41.20	/100.00	&278.3k	/100.00	\\
            L1	&33.41	&10.42	&20.80	/50.49	&176.2k	/63.29	\\
            L2	&33.39	&10.62	&21.00	/50.97	&175.9k	/63.20	\\
            GM	&33.32	&10.35	&20.60	/50.00	&183.8k	/66.03	\\
            GW	&34.24	&10.92	&20.00	/48.54	&168.8k	/60.65	\\
            ES	&33.81	&10.13	&21.00	/50.97	&176.3k	/63.34	\\
            KL	&33.32	&10.62	&21.20	/51.46	&187.5k	/67.35	\\ \midrule
            \multicolumn{5}{c}{ResNet56, CIFAR100} \\ \hline
            Baseline	&27.59	&9.24	&126.80	/100.00	&861.6k	/100.00	\\
            L1	&30.15	&9.34	&63.20	/49.84	&470.8k	/54.64	\\
            L2	&29.48	&9.43	&65.80	/51.89	&513.6k	/59.61	\\
            GM	&29.2	&9.35	&62.30	/49.13	&559.4k	/64.92	\\
            TE	&29.01	&9.33	&65.50	/51.66	&534.3k	/62.01	\\
            ES	&29.49	&9.16	&64.20	/50.63	&554.4k	/64.34	\\
            KL	&29.23	&9.3	&65.10	/51.34	&568.0k	/65.92	\\
            \bottomrule
        \end{tabular}
    \end{center}
    % \vspace{-0.4cm}
    \caption{Benchmarking channel pruning criteria on CIFAR10 and CIFAR100 image classification under the scheme of random pruning.}
    \label{tbl:benchmark_cifar_supp}
    % \vspace{-0.4cm}
\end{table}

\begin{figure*}[!tb]
\begin{center}
    \begin{subfigure}[t]{0.38\textwidth}
    \centering
    \includegraphics[width=0.98\textwidth]{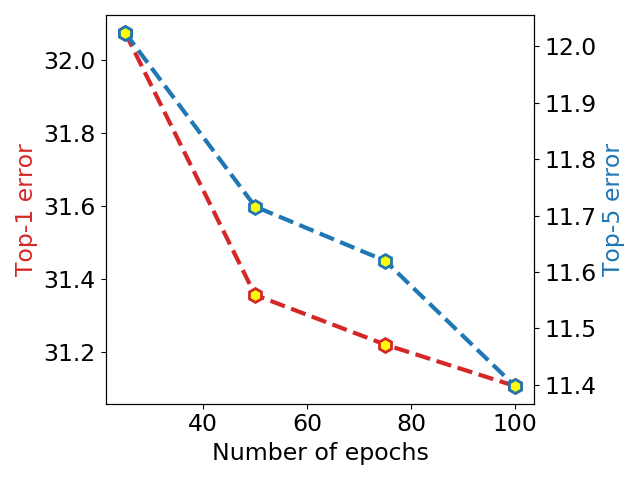}
    \caption{Epochs vs error in ResNet18.}
    \label{fig:resnet18_epoch_supp}
\end{subfigure}
\begin{subfigure}[t]{0.38\textwidth}
    \centering
    \includegraphics[width=0.98\textwidth]{figures/epochs_50.png}
    \caption{Epochs vs. error in ResNet50.}
    \label{fig:resnet50_epoch_supp}
\end{subfigure}
\end{center}
% \vspace{-2pt}
\caption{The influence of the random sample size and fine-tuning epochs on the prediction accuracy.}
\label{fig:ablation_study_supp}
% \vspace{-10pt}
\end{figure*}

\begin{figure*}[!htb]
\begin{subfigure}[t]{0.33\textwidth}
    \centering
    \includegraphics[trim={11 11 11 11},clip,width=0.98\textwidth]{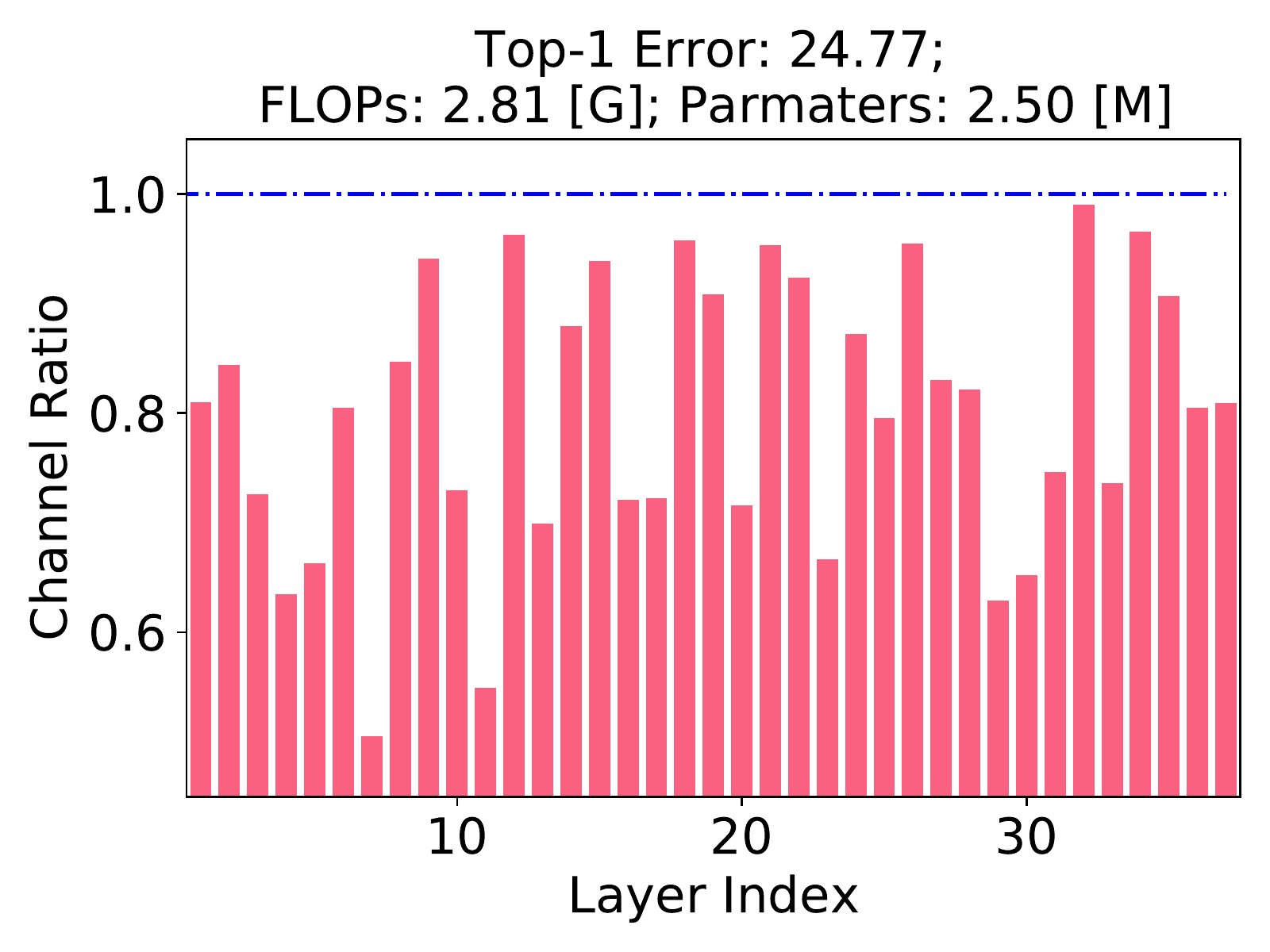}
    \caption{L1.}
    \label{fig:resnet50_channel_ratio_l1}
\end{subfigure}
\begin{subfigure}[t]{0.33\textwidth}
    \centering
    \includegraphics[trim={11 11 11 11},clip,width=0.98\textwidth]{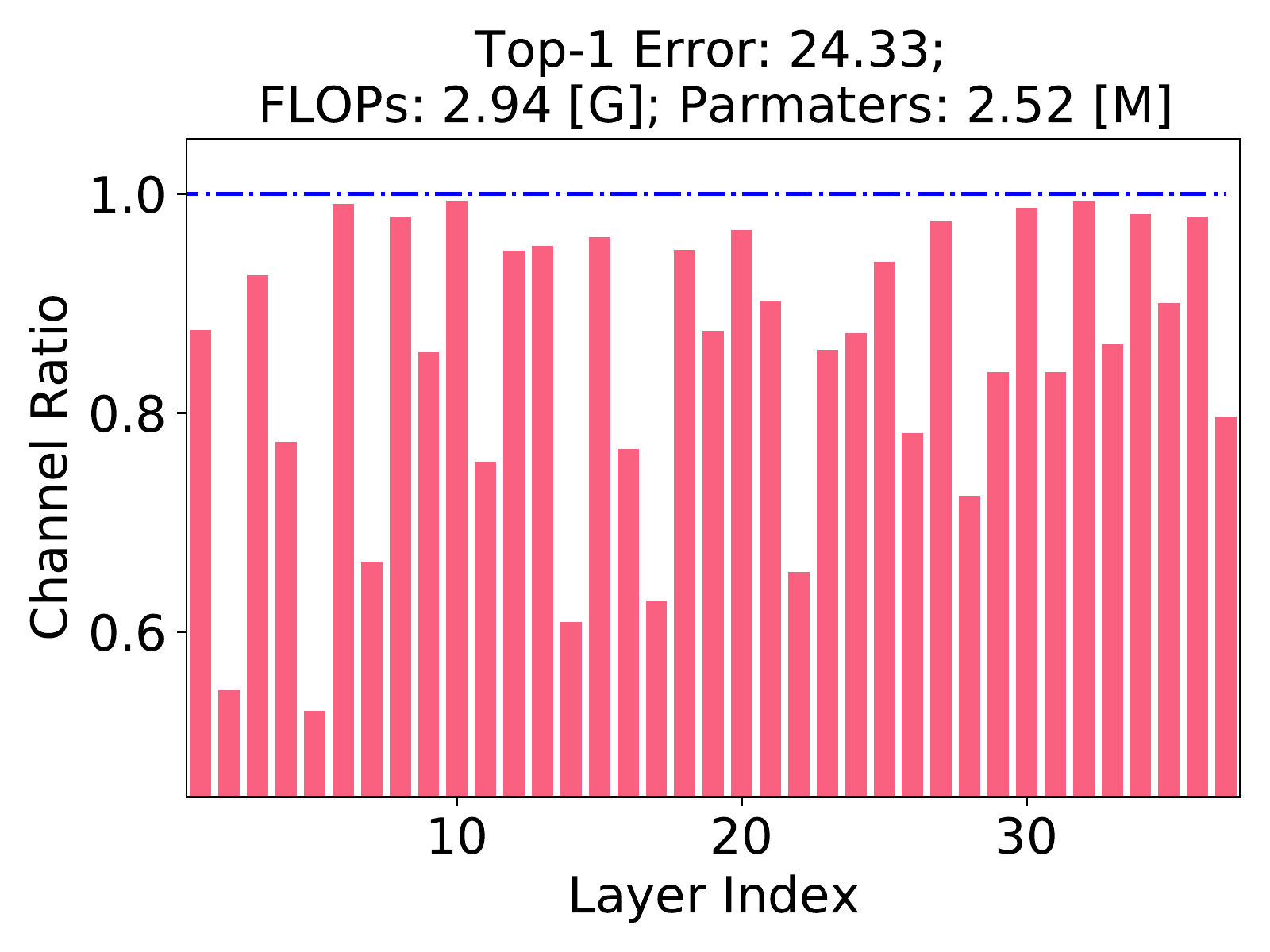}
    \caption{L2.}
    \label{fig:resnet50_channel_ratio_l2}
\end{subfigure}
\begin{subfigure}[t]{0.33\textwidth}
    \centering
    \includegraphics[trim={11 11 11 11},clip,width=0.98\textwidth]{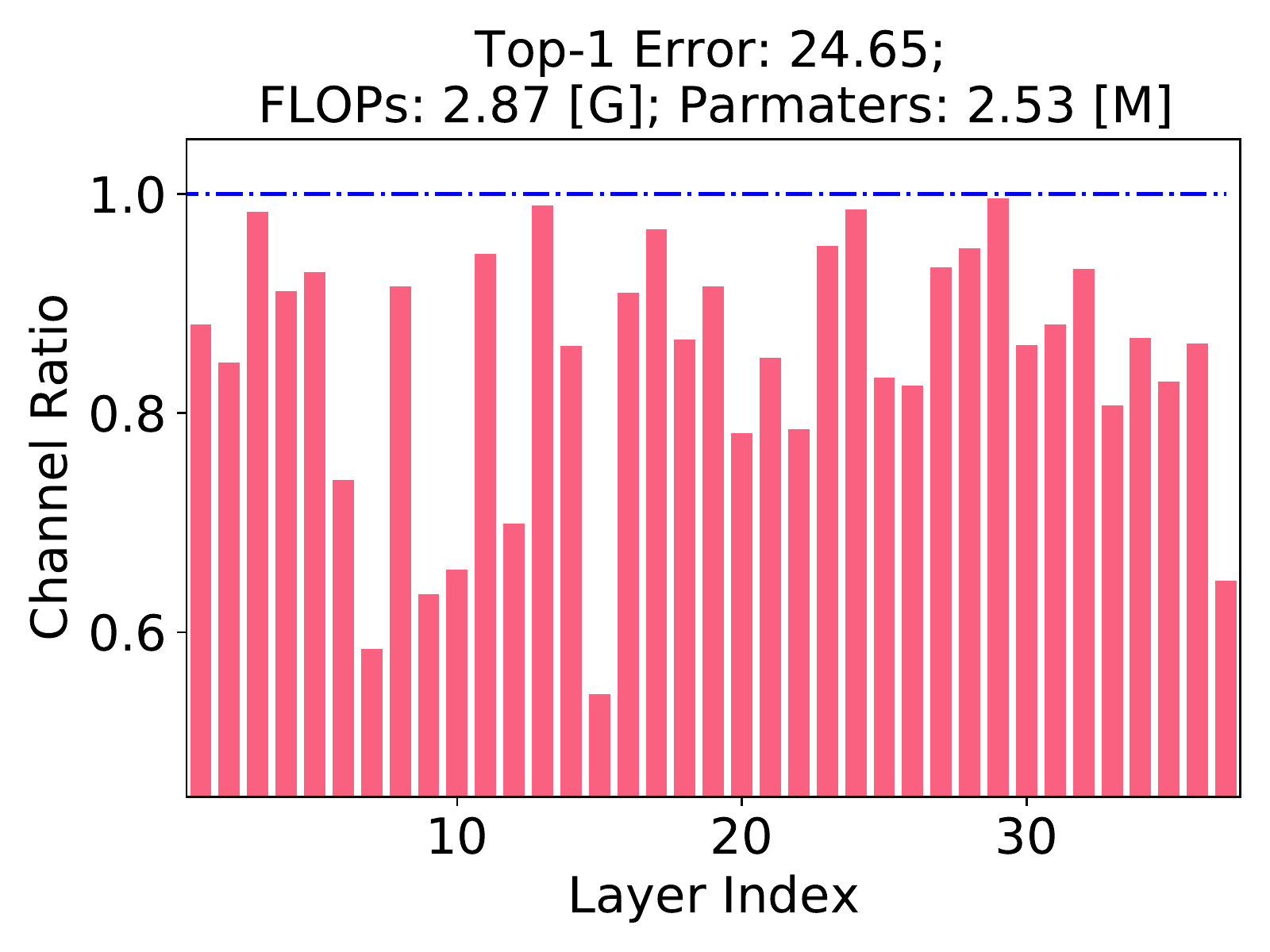}
    \caption{GM.}
    \label{fig:resnet50_channel_ratio_gm}
\end{subfigure}
\begin{subfigure}[t]{0.33\textwidth}
    \centering
    \includegraphics[trim={11 11 11 11},clip,width=0.98\textwidth]{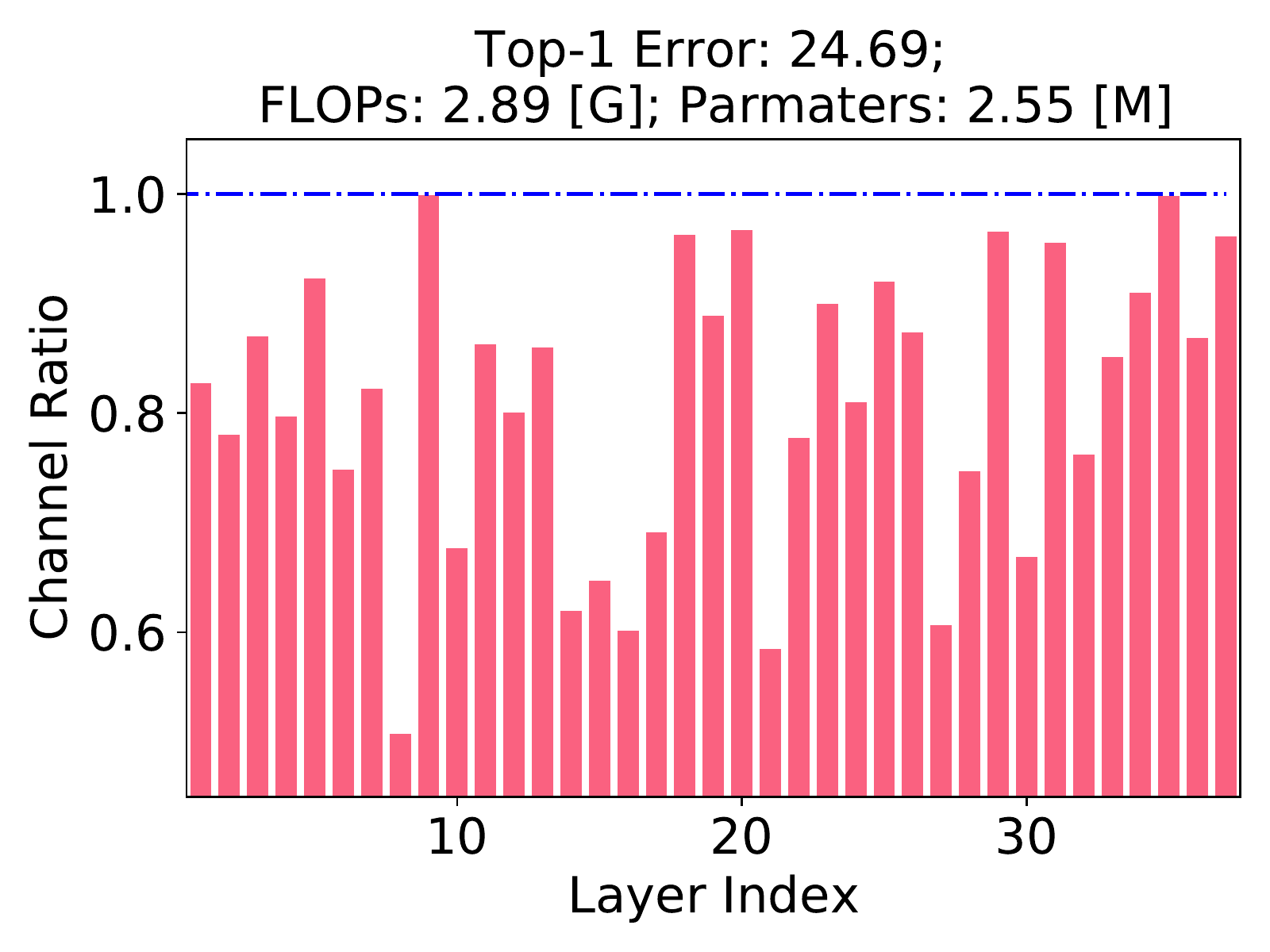}
    \caption{TE.}
    \label{fig:resnet50_channel_ratio_te}
\end{subfigure}
\begin{subfigure}[t]{0.33\textwidth}
    \centering
    \includegraphics[trim={11 11 11 11},clip,width=0.98\textwidth]{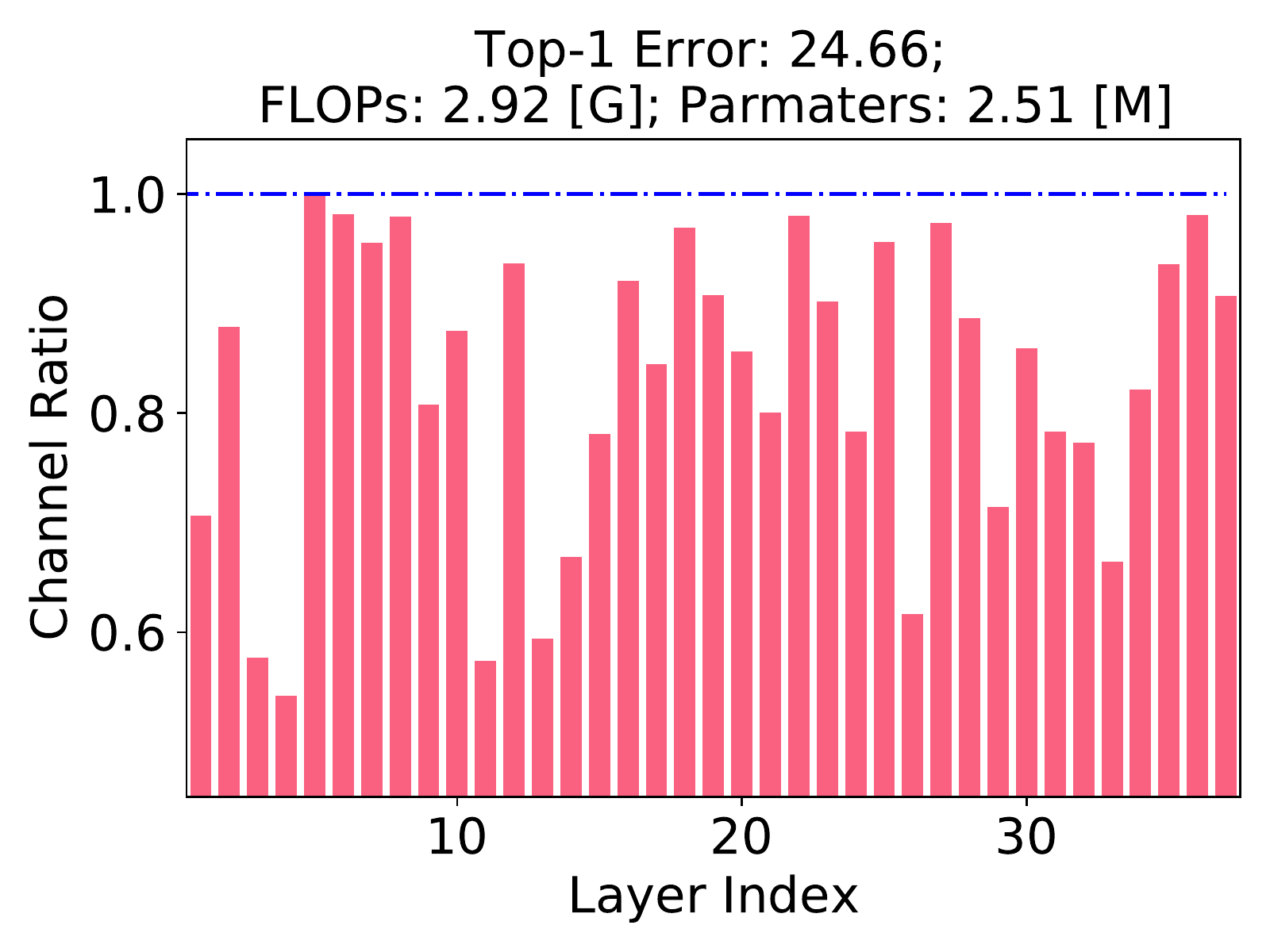}
    \caption{KL.}
    \label{fig:resnet50_channel_ratio_kl}
\end{subfigure}
\begin{subfigure}[t]{0.33\textwidth}
    \centering
    \includegraphics[trim={11 11 11 11},clip,width=0.98\textwidth]{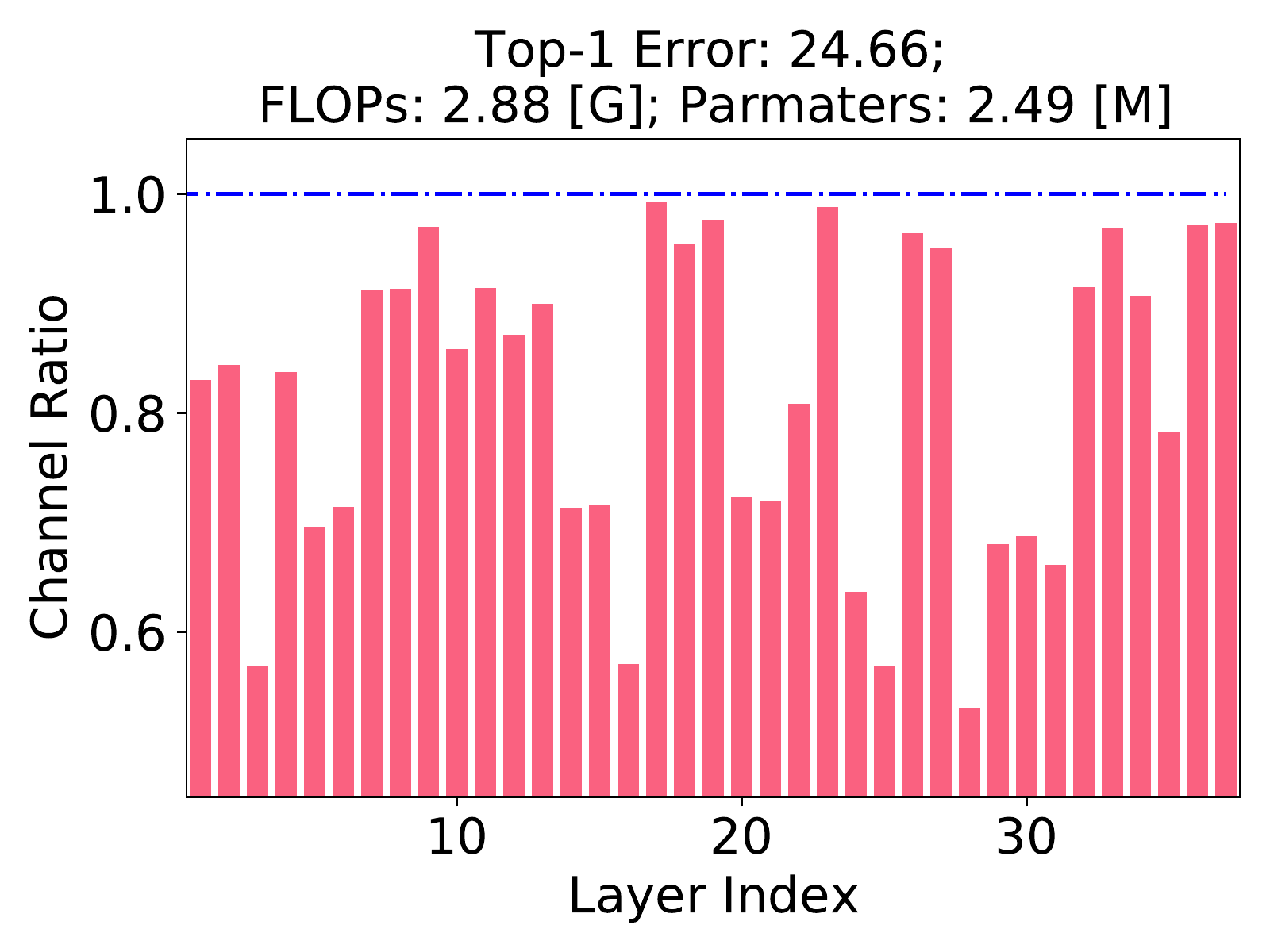}
    \caption{ES.}
    \label{fig:resnet50_channel_ratio_es}
\end{subfigure}
\caption{Percentage of remaining channels of the pruned ResNet50 network. The network pruned by different methods are reported. The pruning ratio is 70\%. The Top-1 error, FLOPs, and number of parameters are also reported in the figure.}
\label{fig:resnet50_channel_ratio}
\end{figure*}

%method -work oblivious
\section{ Upper Bounded Performance of Channel Pruning.} In the this section, we provide the justification of the statement in the main paper ``the performance of the channel pruned network is upper bounded by the original network''. 

In the paper ``The Lottery Ticket Hypothesis'', the authors showed that some pruned networks could learn faster while reaching higher test accuracy and generalizing better than the original one~\cite{frankle2018lottery}.
Yet, the conclusion is derived for unstructured pruning. 
The problems of unstructured pruning and structured pruning are quite different. Unstructured pruning removes single connections in a CNN and results in irregular kernels.
And it is possible that the number of kernels in the resultant sparse network is the same as the original network. 
The capacity of a network could be fully utilized by the sparse network.
This is why unstructured pruning could easily lead to an extremely pruned network without accuracy drop while for structured pruning researchers struggle with the trade-off between accuracy drop and compression ratio.
Without expanding the search space (\ie changing the position of pooling layers~\cite{liu2019metapruning}, widening the network~\cite{yu2018slimmable,yu2019autoslim,li2021heterogeneity}), it is very difficult to find a pruned network with better performance. Thus, we can safely conclude that the performance of channel pruned networks is upper bounded by the original networks.

\section{Pruning Residual Blocks}
\label{sec:pruning_residual}

Pruning a normal convolutional layer is straightforward. But when it comes to the residual blocks in MobileNetV2~\cite{sandler2018mobilenetv2} and ResNet~\cite{he2016deep}, some special measures should be taken. For the residual blocks in MobileNetV2 and ResNet, there is a skip connection that adds the input of the block to the output of the block so that the block learns a residual component. Since the input and output of residual blocks are connected, the number of output channels of several residual blocks are the same. When pruning the residual block, their output channels should be pruned together. For both of the pruning settings explained in the main paper, \ie pruning pre-trained network and and pruning from scratch, we set the same pruning ratio for the convolutional layers that are connected by skip connection. 

Special treatments should also be taken when computing the importance score according to different pruning criteria. \textbf{I. L1/L2/GM.} For the convolutional layers that are connected by skip connection, their individual importance scores are first computed and then added up. The summation result is used as the final importance score. \textbf{II. TE.} As in the original paper, gates with weights equal to 1 and dimensionality equal to the number of output channels are append to the Batch Normalization layers. The importance score are first computed based on the gates and then added for the layers that are skip-connected. \textbf{III. ES.} The maximum empirical sensitivity is computed for layers that are connected by skip connections. \textbf{IV. KL.} To compute the KL divergence for the output probability of the pruned and original networks, masks that selects the output channels should be added to the convolutional layers. For the convolutional layers that are skip-connected,  we set the same mask for them so that the same KL divergence score can be computed for all of them.

\section{More Experimental Results}
\label{sec:more_results}

More experimental results are shown in this section. The results for CIFAR image classification are summarized in Table~\ref{tbl:benchmark_cifar_supp}. Besides the results in the main paper, results of ResNet20 on CIFAR10 and ResNet56 on CIFAR100  are also included. As in the main paper, a couple of pruning criteria are compared including the traditional L1 and L2 norm of the filters (L1, L2), and the recent method based on geometric median (GM)~\cite{he2019filter}, Taylor expansion (TE)~\cite{molchanov2019importance}, KL-divergence importance metric (KL)~\cite{luo2020neural} and empirical sensitivity analysis (ES)~\cite{liebenwein2019provable}. The additional results strengthen the conclusion in the main paper. That is, under the scheme of random pruning, the pruning criteria for selecting different channels are less important.

The influence of fine-tuning epochs on the final accuracy of the pruned network is shown in Fig.~\ref{fig:ablation_study_supp}. The result for ResNet-50 is shown in Fig.~\ref{fig:resnet50_epoch_supp}. The result for ResNet-18 is shown in Fig.~\ref{fig:resnet18_epoch_supp}. When the number of fine-tuning epochs is increased from 25 to 100, the Top-1 and Top-5 error of ResNet-50 drops by 0.75\% and 0.4\%, respectively. For ResNet-18, the Top-1 error rate and Top-5 error rate drop by 0.97\% and 0.62\%, respectively. This shows the significant influence of fine-tuning epochs. 

\begin{figure*}[!htb]
\begin{subfigure}[t]{0.33\textwidth}
    \centering
    \includegraphics[trim={6 0 11 11},clip,width=0.98\textwidth]{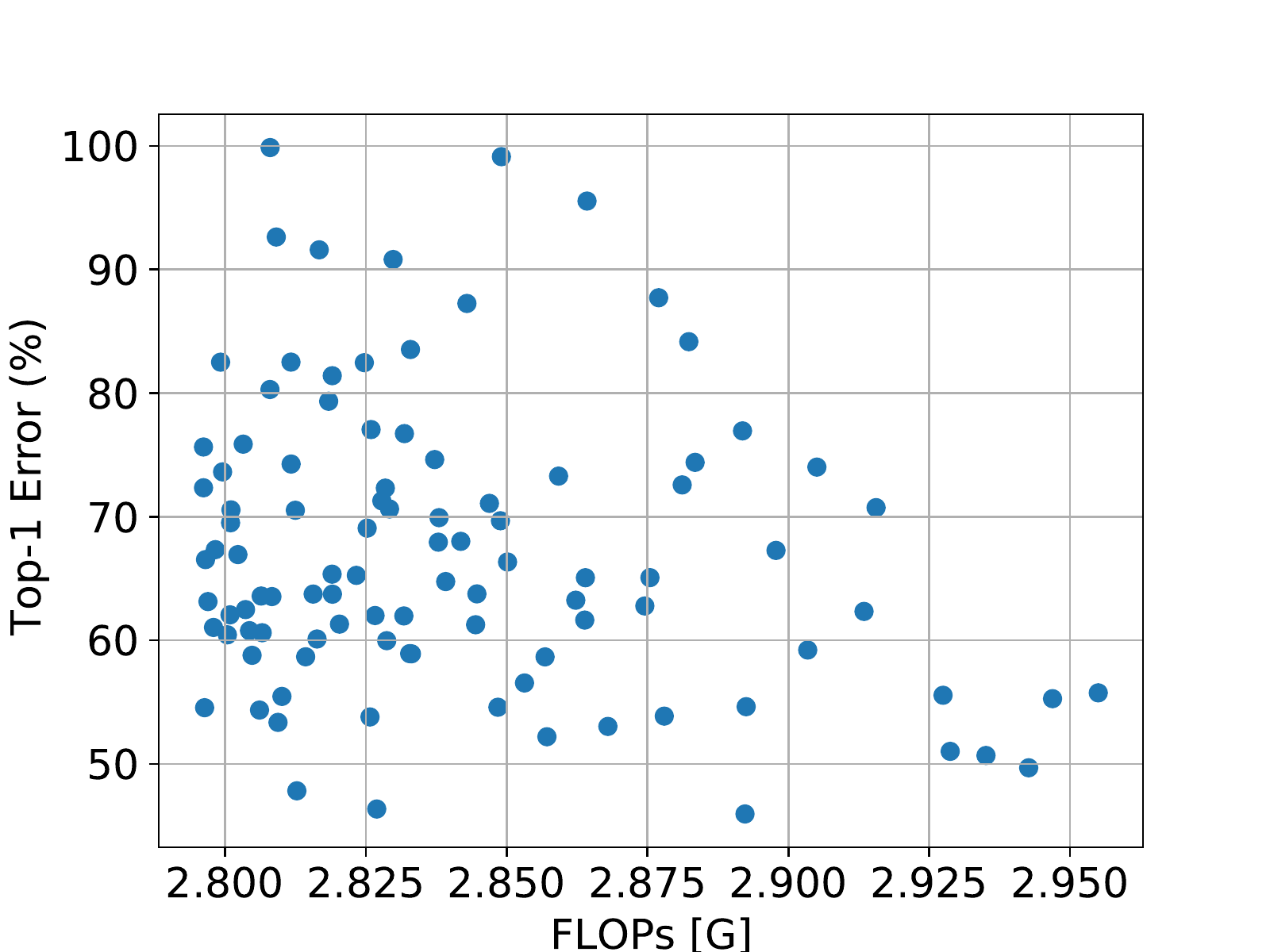}
    \caption{L1.}
    \label{fig:resnet50_flops_dist_l1}
\end{subfigure}
\begin{subfigure}[t]{0.33\textwidth}
    \centering
    \includegraphics[trim={6 0 11 11},clip,width=0.98\textwidth]{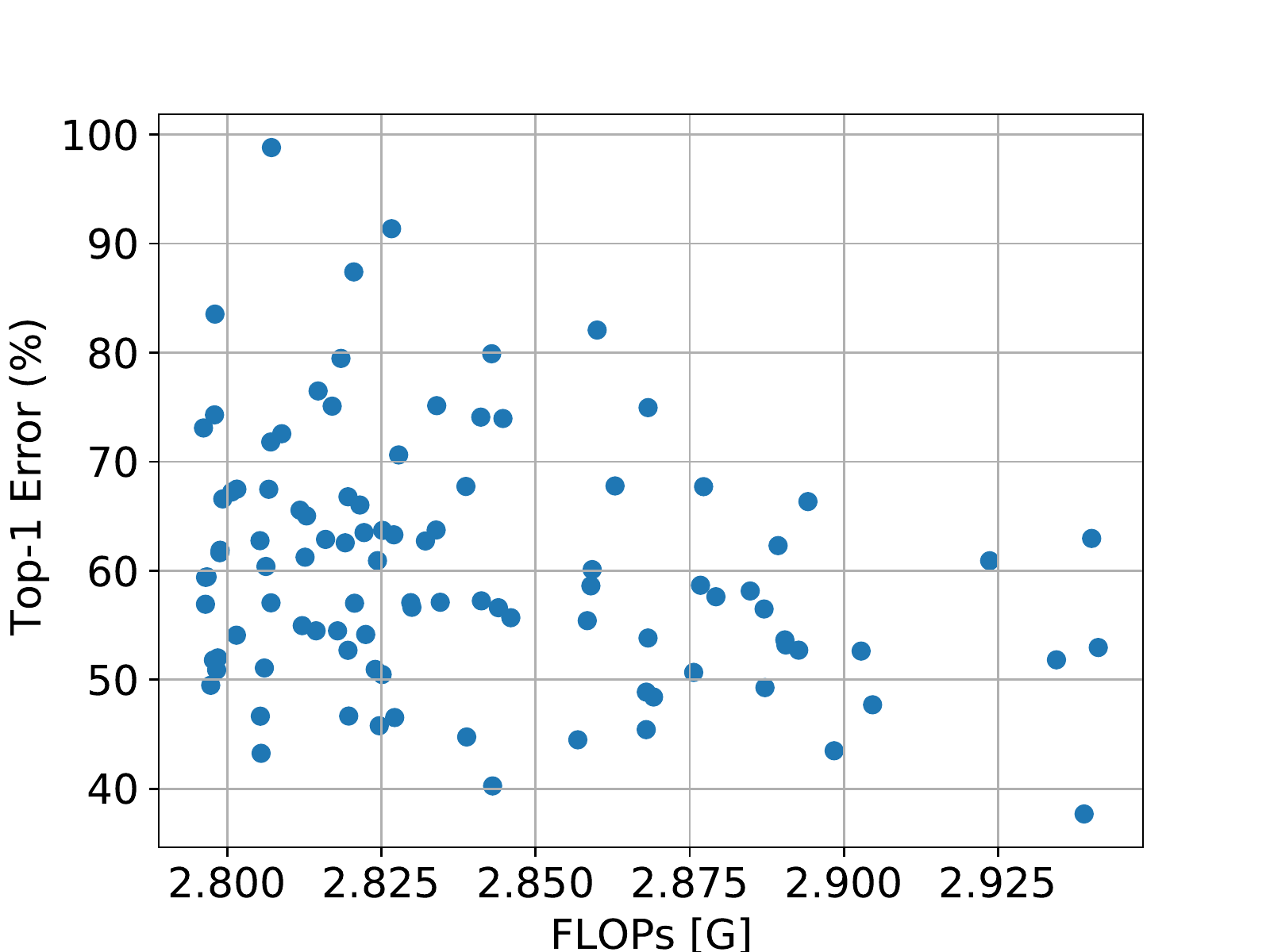}
    \caption{L2.}
    \label{fig:resnet50_flops_dist_l2}
\end{subfigure}
\begin{subfigure}[t]{0.33\textwidth}
    \centering
    \includegraphics[trim={6 0 11 11},clip,width=0.98\textwidth]{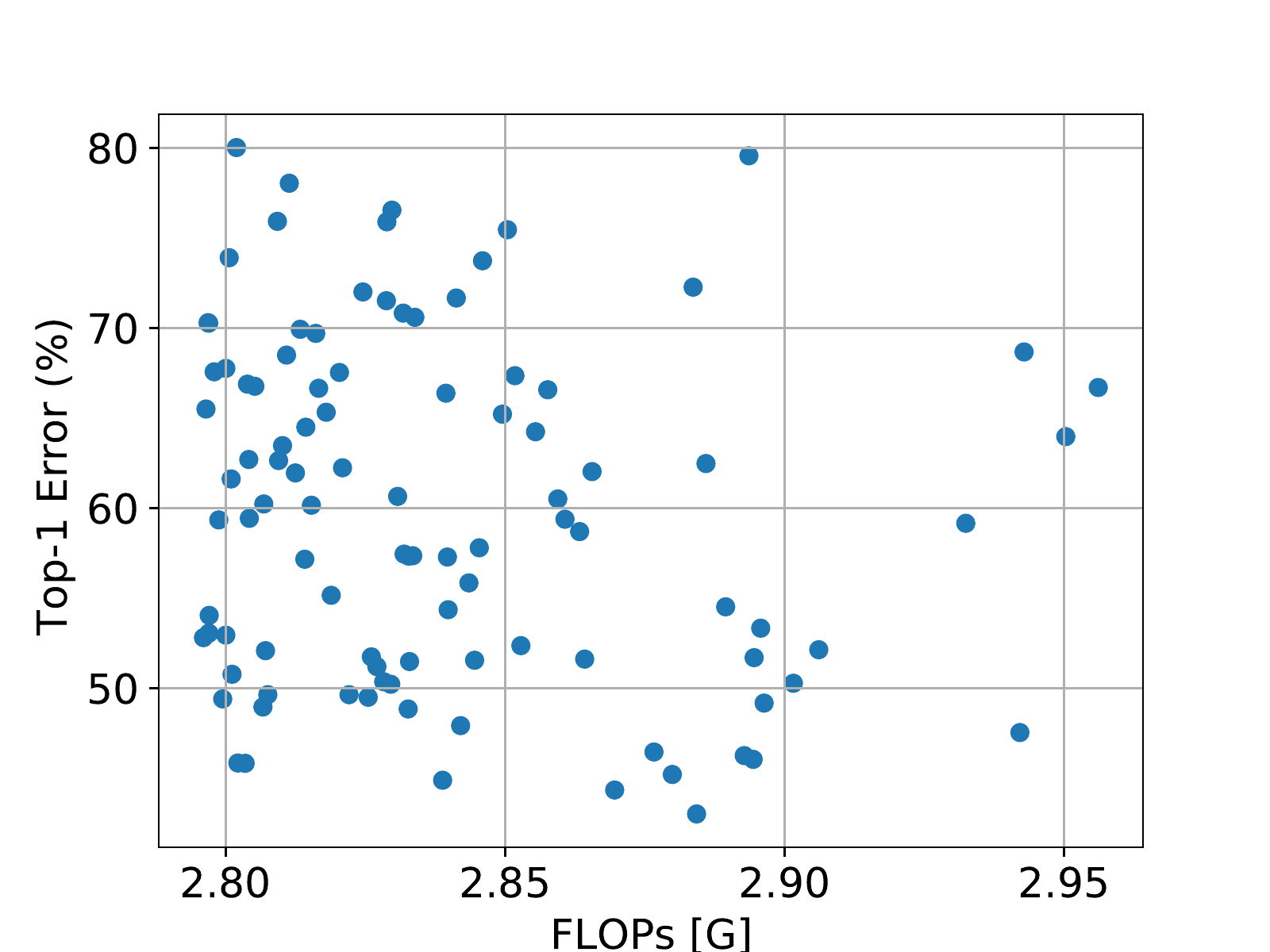}
    \caption{GM.}
    \label{fig:resnet50_flops_dist_gm}
\end{subfigure}
\begin{subfigure}[t]{0.33\textwidth}
    \centering
    \includegraphics[trim={6 0 11 11},clip,width=0.98\textwidth]{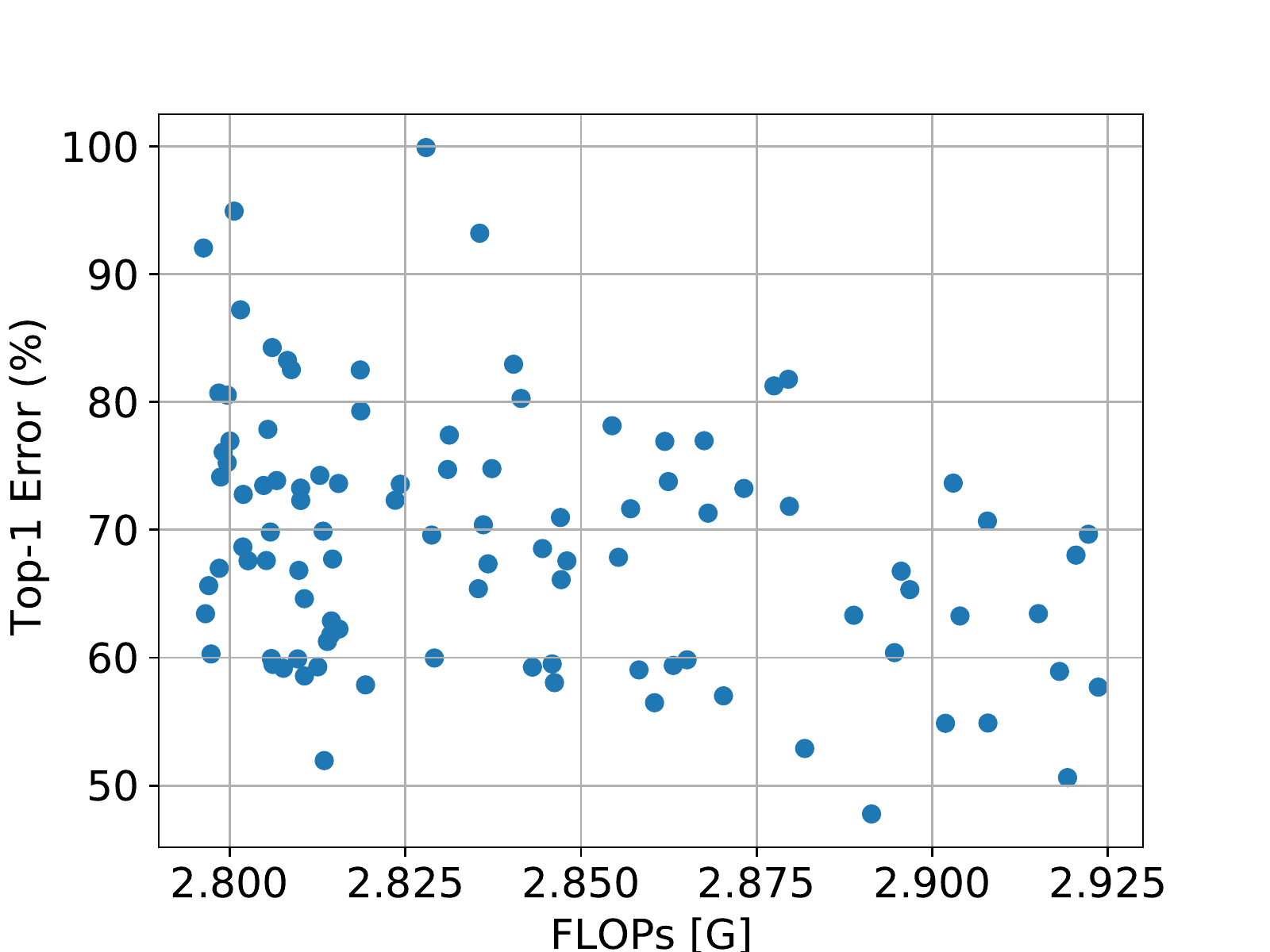}
    \caption{TE.}
    \label{fig:resnet50_flops_dist_te}
\end{subfigure}
\begin{subfigure}[t]{0.33\textwidth}
    \centering
    \includegraphics[trim={6 0 11 11},clip,width=0.98\textwidth]{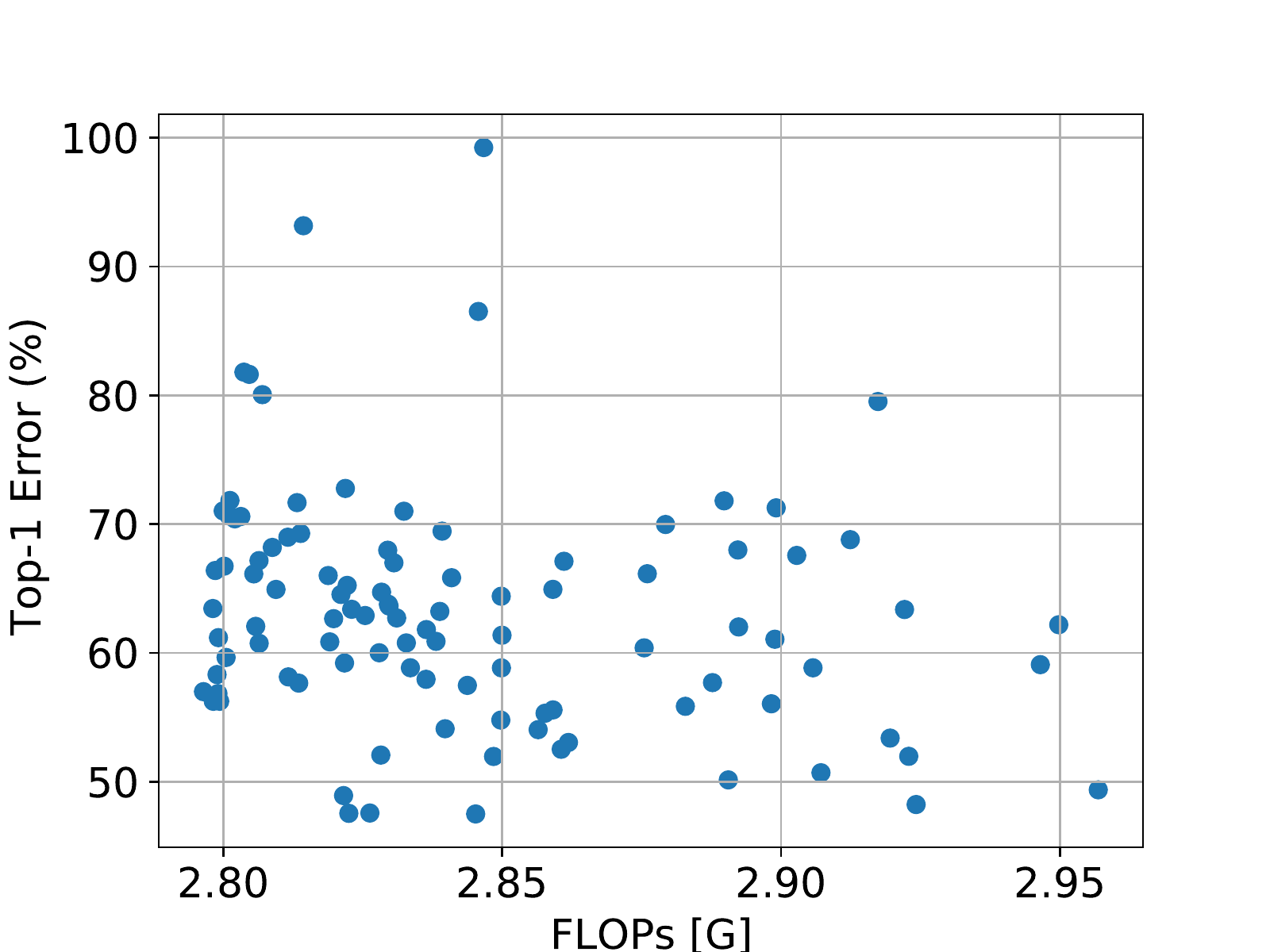}
    \caption{KL.}
    \label{fig:resnet50_flops_dist_kl}
\end{subfigure}
\begin{subfigure}[t]{0.33\textwidth}
    \centering
    \includegraphics[trim={6 0 11 11},clip,width=0.98\textwidth]{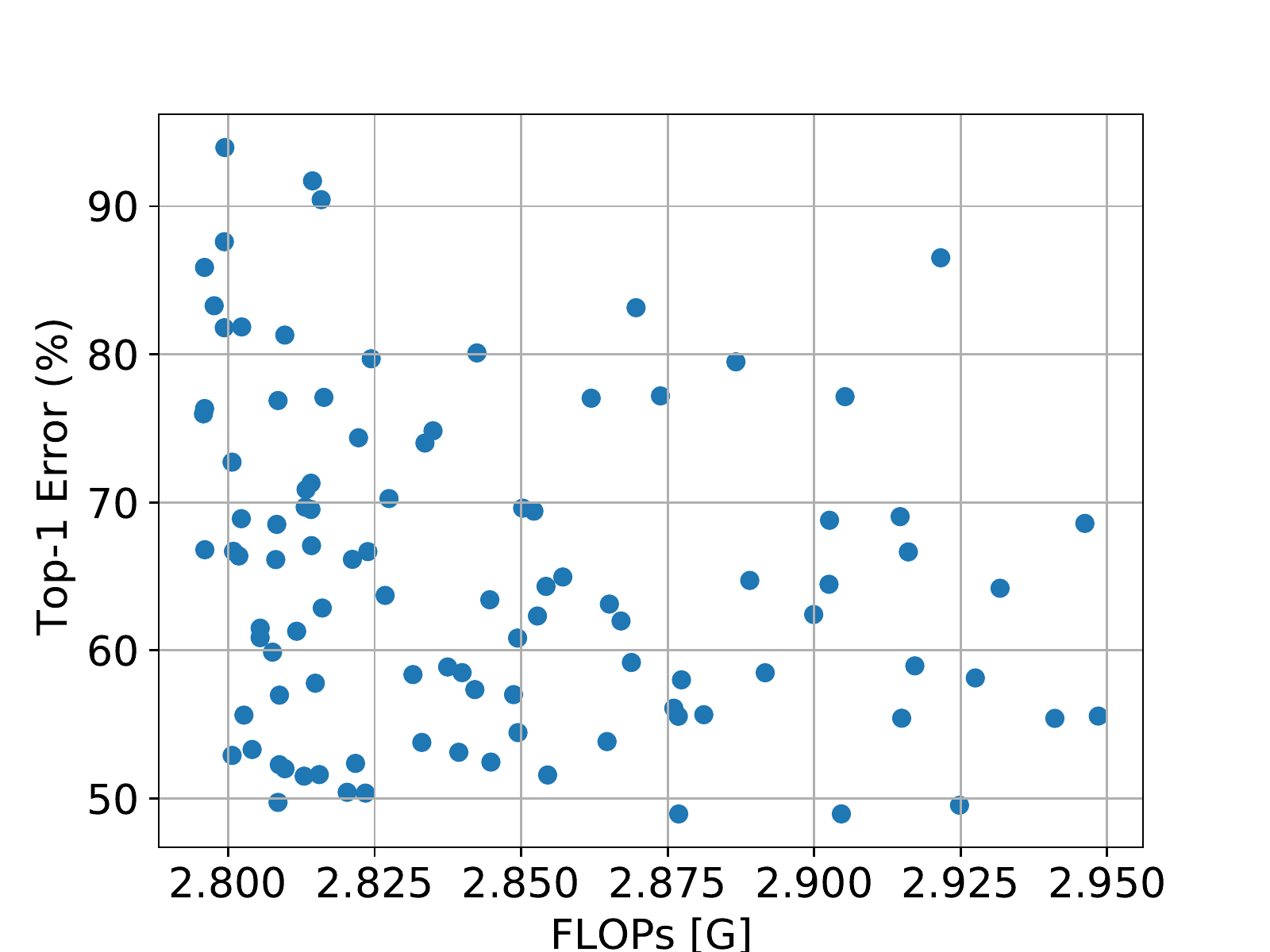}
    \caption{ES.}
    \label{fig:resnet50_flops_dist_es}
\end{subfigure}
\caption{Accuracy distribution of network samples with respect to FLOPs for different pruning criteria. The original network is ResNet50 trained for ImageNet classification. The network pruning ratio is 70\%.}
\label{fig:resnet50_flops_dist}
\end{figure*}

\begin{figure*}[!htb]
\begin{subfigure}[t]{0.33\textwidth}
    \centering
    \includegraphics[trim={6 0 11 11},clip,width=0.98\textwidth]{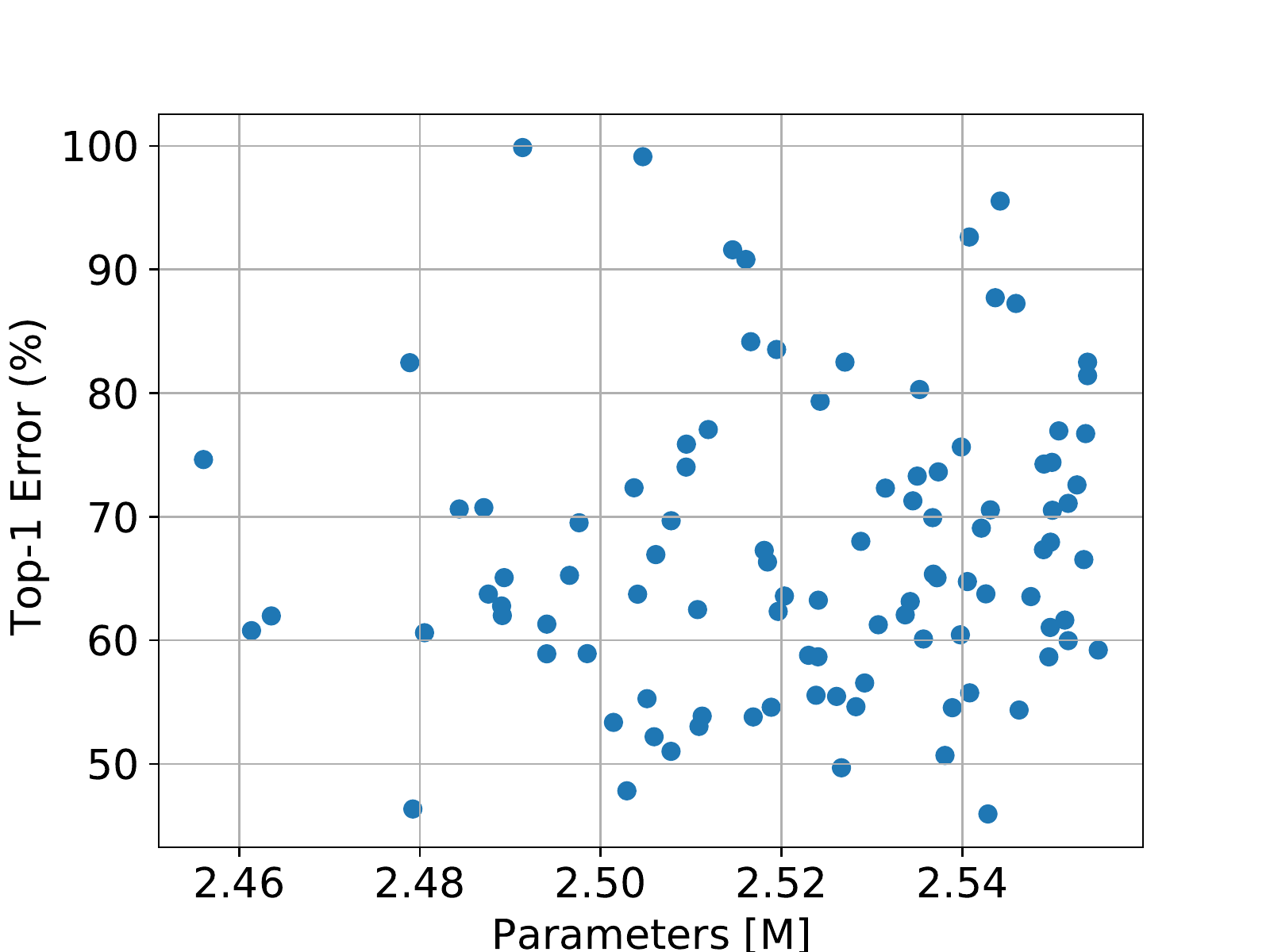}
    \caption{L1.}
    \label{fig:resnet50_params_dist_l1}
\end{subfigure}
\begin{subfigure}[t]{0.33\textwidth}
    \centering
    \includegraphics[trim={6 0 11 11},clip,width=0.98\textwidth]{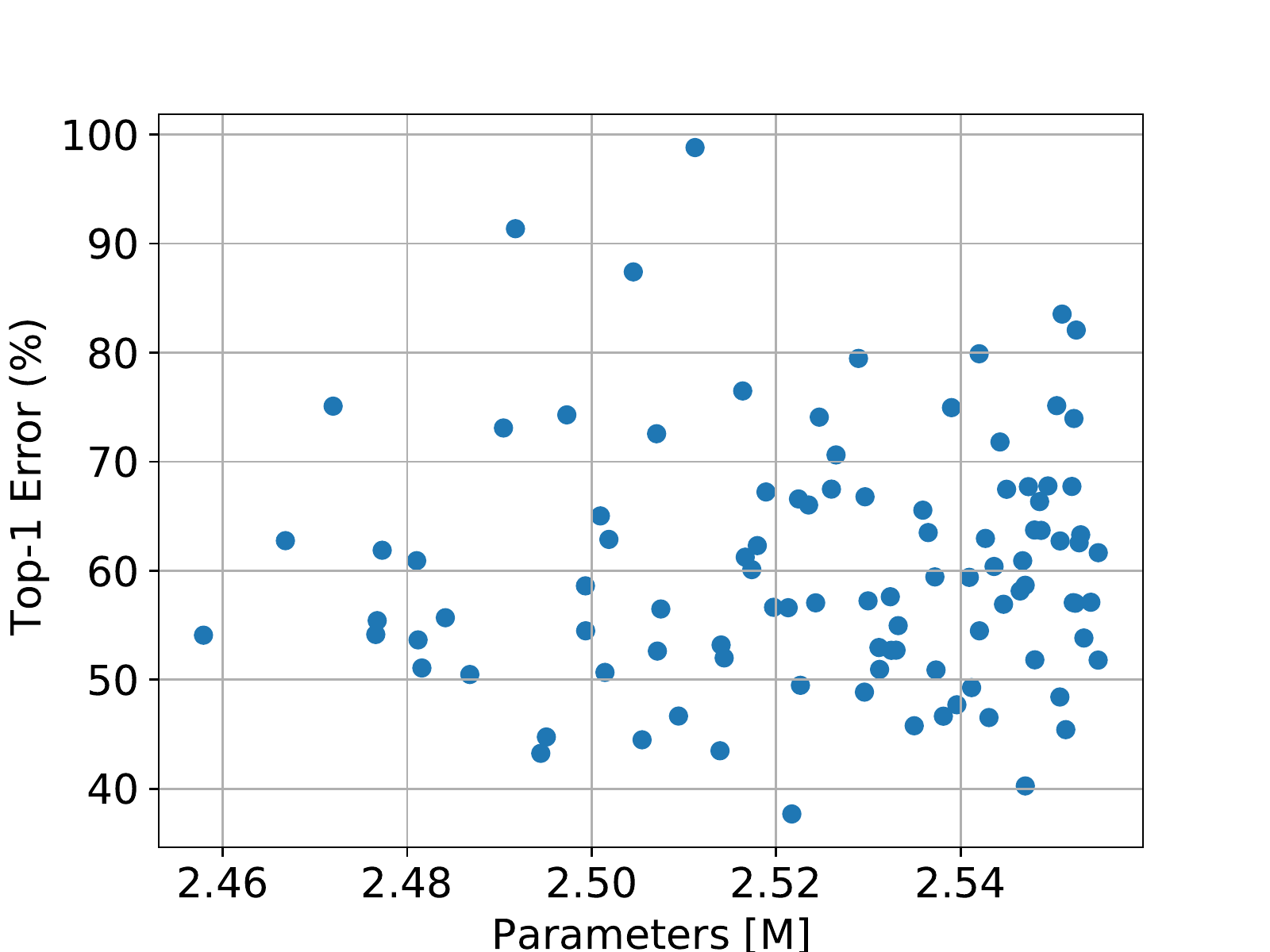}
    \caption{L2.}
    \label{fig:resnet50_params_dist_l2}
\end{subfigure}
\begin{subfigure}[t]{0.33\textwidth}
    \centering
    \includegraphics[trim={6 0 11 11},clip,width=0.98\textwidth]{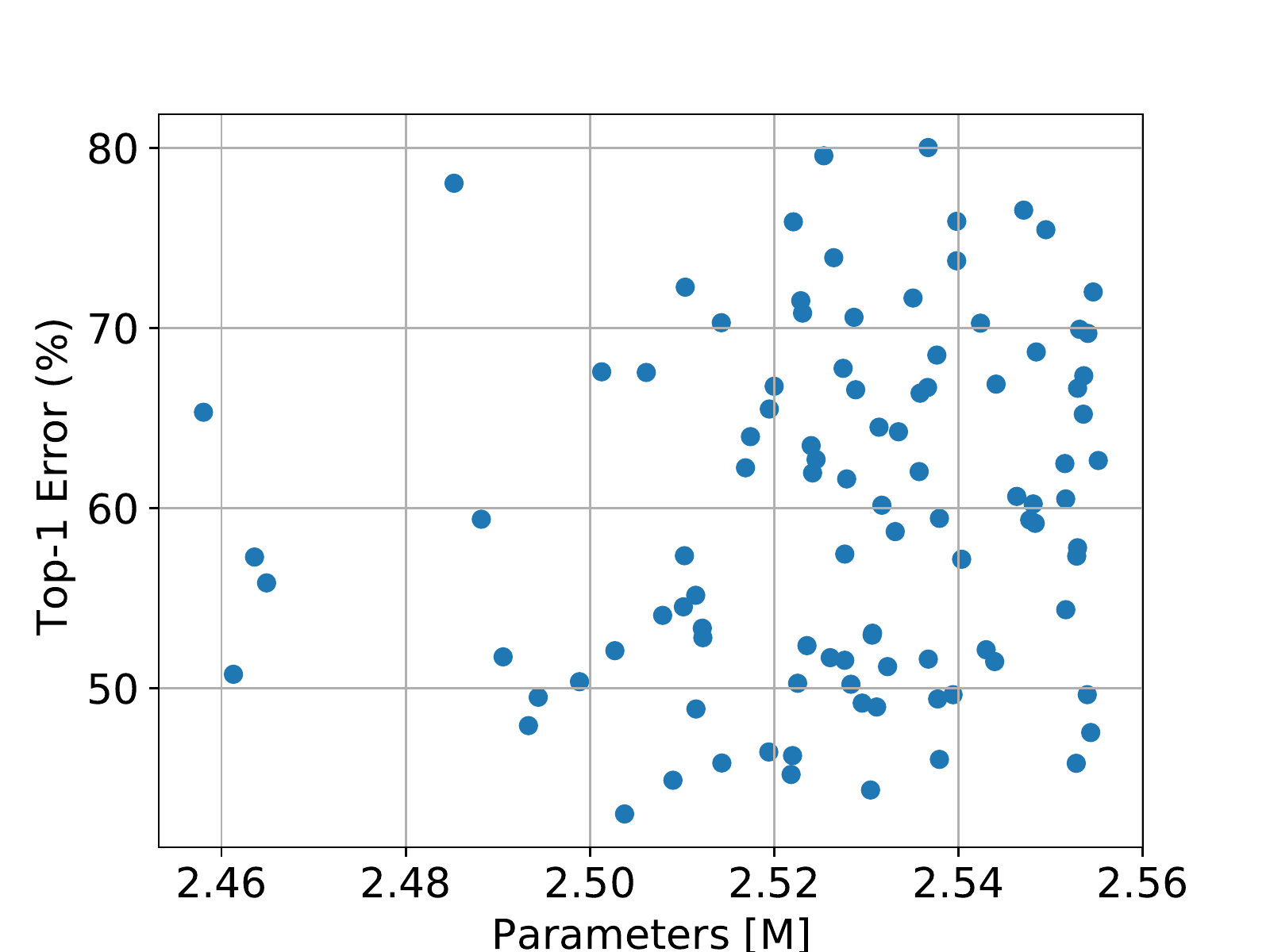}
    \caption{GM.}
    \label{fig:resnet50_params_dist_gm}
\end{subfigure}
\begin{subfigure}[t]{0.33\textwidth}
    \centering
    \includegraphics[trim={6 0 11 11},clip,width=0.98\textwidth]{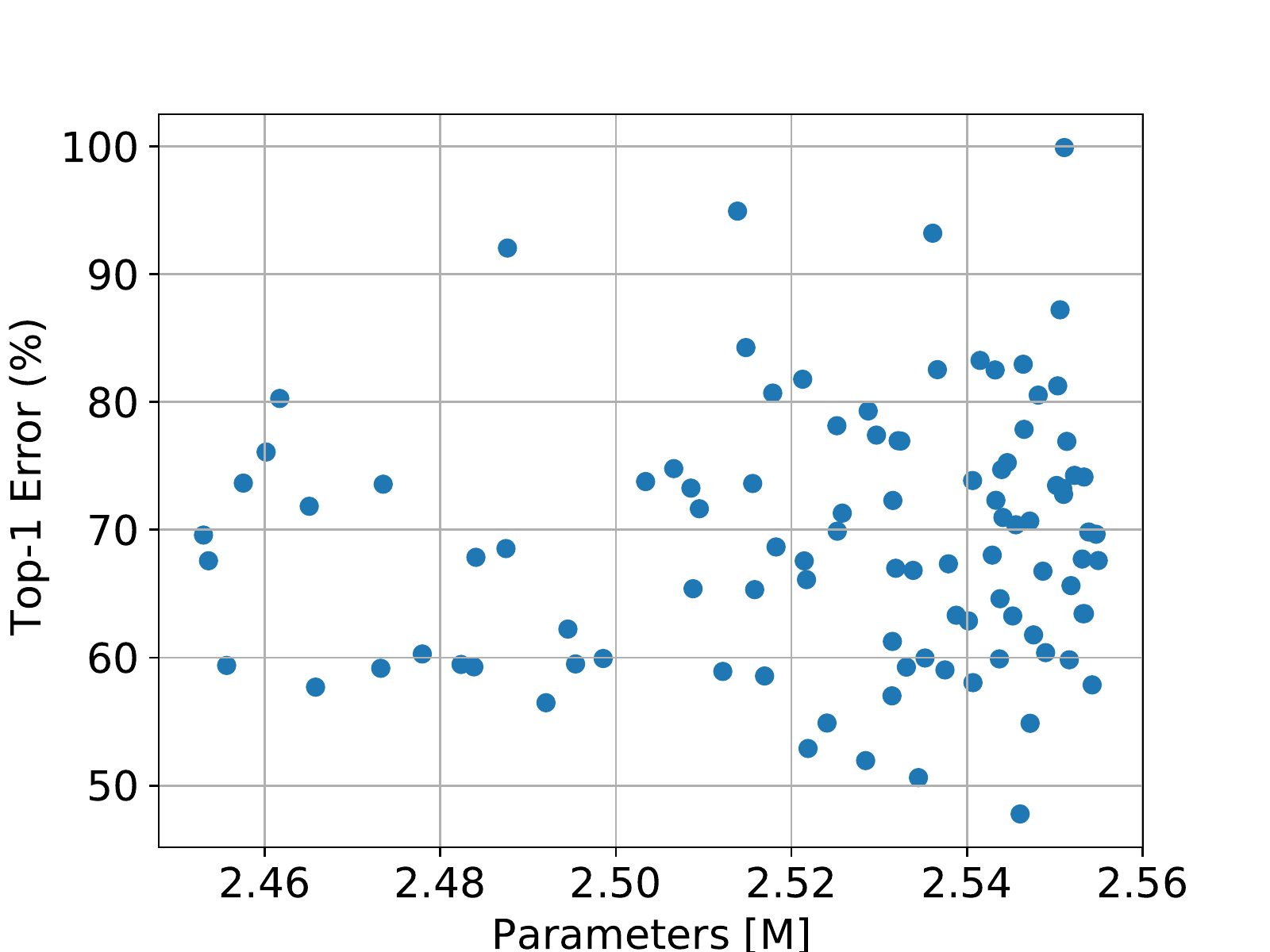}
    \caption{TE.}
    \label{fig:resnet50_params_dist_te}
\end{subfigure}
\begin{subfigure}[t]{0.33\textwidth}
    \centering
    \includegraphics[trim={6 0 11 11},clip,width=0.98\textwidth]{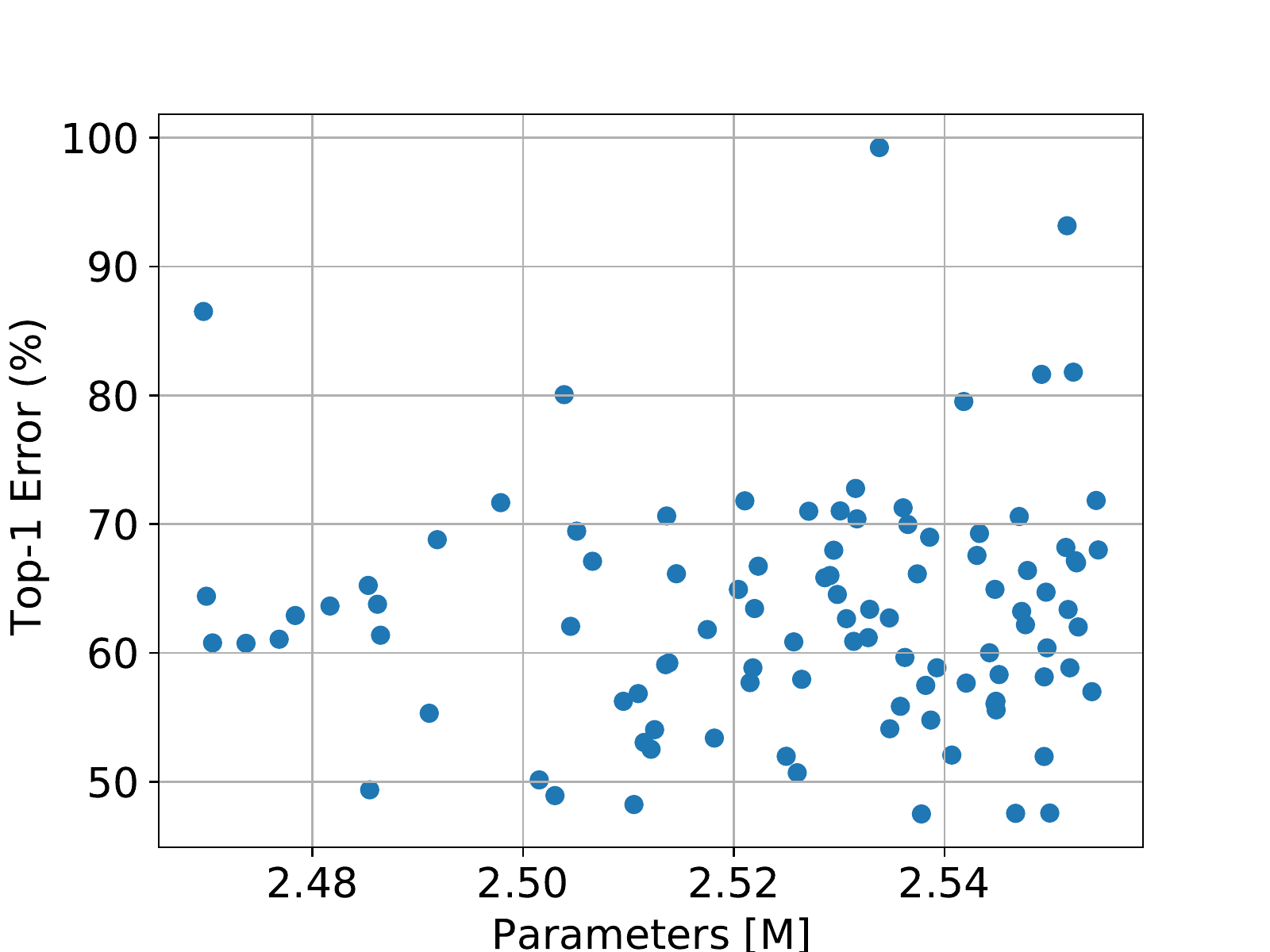}
    \caption{KL.}
    \label{fig:resnet50_params_dist_kl}
\end{subfigure}
\begin{subfigure}[t]{0.33\textwidth}
    \centering
    \includegraphics[trim={6 0 11 11},clip,width=0.98\textwidth]{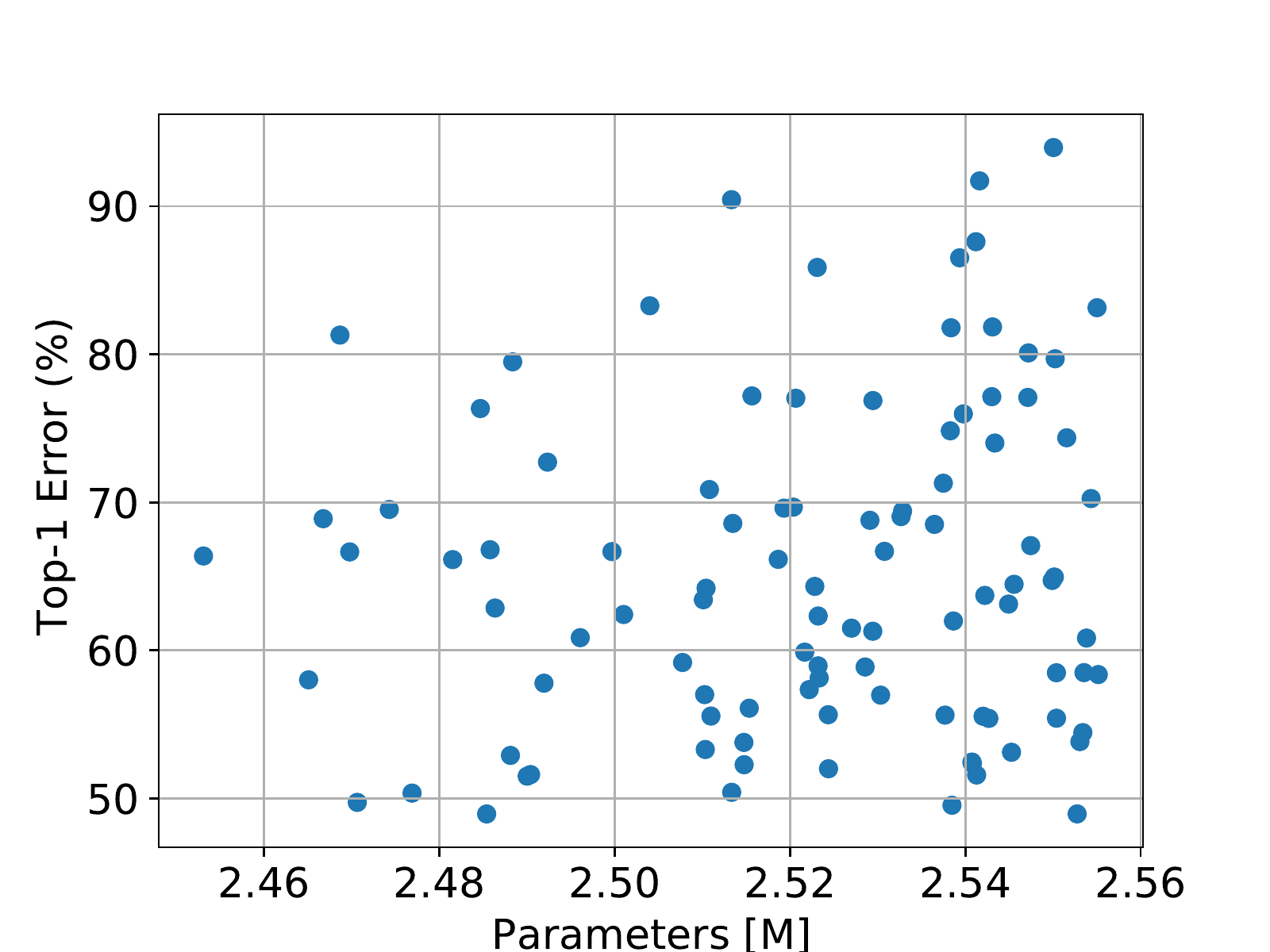}
    \caption{ES.}
    \label{fig:resnet50_params_dist_es}
\end{subfigure}
\caption{Accuracy distribution of network samples with respect to the number of parameters for different pruning criteria. The original network is ResNet50 trained for ImageNet classification. The network pruning ratio is 70\%.}
\label{fig:resnet50_params_dist}
\end{figure*}

In Fig.~\ref{fig:resnet50_channel_ratio}, the ratio of remaining channels for each of the convolutional layer is plotted. The original network is ResNet50 for ImageNet classification and the overall pruning ratio is 70\%. The Top-1 error, FLOPs, and number of parameters are also reported in the figure. In Fig.~\ref{fig:resnet50_flops_dist}, the accuracy distribution of the random pruned networks with respect to FLOPs is shown. Note that the networks are only updated by minimizing the squared difference between the features maps of the pruned and original network. Fine-tuning has not been conducted during this step. As can be seen, both good sub-networks with low error rate and less accurate sub-networks can be sampled. And the aim is to search the sub-networks with higher accuracy. Similar to Fig.~\ref{fig:resnet50_flops_dist}, the accuracy distribution with respect to the number of parameters is shown in Fig~\ref{fig:resnet50_params_dist}.

